%% file: main.tex
\documentclass{article}

\usepackage{iclr2027_conference,times}
\usepackage{microtype}
\usepackage{graphicx}
\usepackage{booktabs}
\usepackage{hyperref}
\usepackage{url}
\usepackage{placeins}

\usepackage{amsmath,amssymb,mathtools}
\usepackage{xspace}
\usepackage{tikz}
\usetikzlibrary{arrows.meta,calc,fit,positioning}

\newcommand{\method}{\textsc{PermVLA}\xspace}
\newcommand{\caploss}{\mathcal{L}_{\mathrm{CAP}}}
\newcommand{\ltrloss}{\mathcal{L}_{\mathrm{LTR}}}
\newcommand{\context}{\mathcal{C}}
\newcommand{\actions}{\mathcal{A}}
\newcommand{\tablebodyfont}{\footnotesize}

\title{PermVLA: Factorization Order as a Regularizer for VLA Learning}

\input{authors}

\iclrfinalcopy
\hypersetup{
  hidelinks,
  pdftitle={PermVLA: Factorization Order as a Regularizer for VLA Learning},
  pdfauthor={Yanqiao Chen, Yuhan Rui, Dongsheng Hou, Zijie Nie, Yutong Wan, Qi Hao}
}

\begin{document}

\maketitle
\lhead{arXiv preprint}
\begingroup
\renewcommand{\thefootnote}{\fnsymbol{footnote}}
\footnotetext[1]{Equal contribution.}
\endgroup
\setcounter{footnote}{0}

\begin{abstract}
Vision-language-action (VLA) policies commonly learn action chunks through a
fixed left-to-right (LTR) factorization, although the same expert trajectory
distribution admits many valid chain-rule factorizations.  We identify
factorization order as an overlooked regularization choice and introduce
causally anchored permutation (CAP), which samples action reveal orders with
a tunable chronological prefix.  Its auxiliary
objective trains one shared policy to predict actions from different known
subsets of the same expert chunk, while deployment retains deterministic LTR
control.  We call this \emph{conditional-set augmentation}: it creates multiple
conditional prediction problems from one expert chunk without adding
demonstrations.  This discourages reliance on the single chronological prefix
used by ordinary teacher forcing.  Controlled experiments show that CAP
consistently outperforms standard LTR training on LIBERO and LIBERO-Plus, with
the same advantage appearing in cross-dataset CALVIN evaluation.  A diagnostic
that measures the expected squared difference between a chunk's joint log
likelihood under two reveal orders verifies that CAP training internalizes
agreement across reveal orders.  These findings position sampled
subset-conditioned auxiliary objectives as a general recipe for constructing
VLA regularizers, illustrated by an extension to diffusion action generators.
\end{abstract}

\input{sections/01_introduction}
\input{sections/02_related_work}
\input{sections/03_method}
\input{sections/04_experiments}
\input{sections/05_discussion}
\input{sections/06_conclusion}

\bibliography{references}
\bibliographystyle{iclr2027_conference}

\clearpage
\FloatBarrier
\appendix
\input{sections/appendix_a_experimental_details}
\input{sections/appendix_b_permdiffusion}
\input{sections/appendix_c_theory}
\input{sections/appendix_d_robot_visualization}

\end{document}

%% file: authors.tex
\author{%
\textbf{Yanqiao Chen\textsuperscript{1,*},
Yuhan Rui\textsuperscript{1,*},
Dongsheng Hou\textsuperscript{1,*}} \\
\textbf{Zijie Nie\textsuperscript{2},
Yutong Wan\textsuperscript{1},
Qi Hao\textsuperscript{1}} \\[0.5em]
\normalfont\small\textsuperscript{1}Southern University of Science and Technology \\
\normalfont\small\textsuperscript{2}Huazhong University of Science and Technology
}

%% file: sections/01_introduction.tex
\section{Introduction}
\label{sec:introduction}

Vision-language-action (VLA) systems learn a conditional joint distribution
over a chunk of future actions.  Recent work has explored robot pretraining,
action tokenization, diffusion and flow objectives, post-training, and
asynchronous execution
\citep{kim2024openvla,kim2025oft,black2024pi0,pi2025pi05,shukor2025smolvla}.
The chain rule introduces another modeling choice: the order in which action
variables are revealed and predicted.  Autoregressive VLA training usually
fixes this order to chronological action time, conventionally modeled
left-to-right (LTR), leaving factorization order hidden inside the action
objective.  We study it
as an independent modeling choice and ask whether alternative valid
factorizations can regularize VLA training.

Under LTR training, each action slot is always predicted from the same family
of preceding physical actions.  Complete expert chunks make many other valid
conditioning subsets available.  We hypothesize that exclusive training on
one teacher-forced chain encourages the policy to rely on
that single reveal order.  Predicting the same physical action from different
known subsets should instead favor conditional dependencies among action steps
that remain predictive across conditioning patterns.  This may improve
closed-loop control when visual, linguistic, or embodiment context shifts.

\begin{figure*}[t]
  \centering
  \includegraphics[width=1\linewidth]{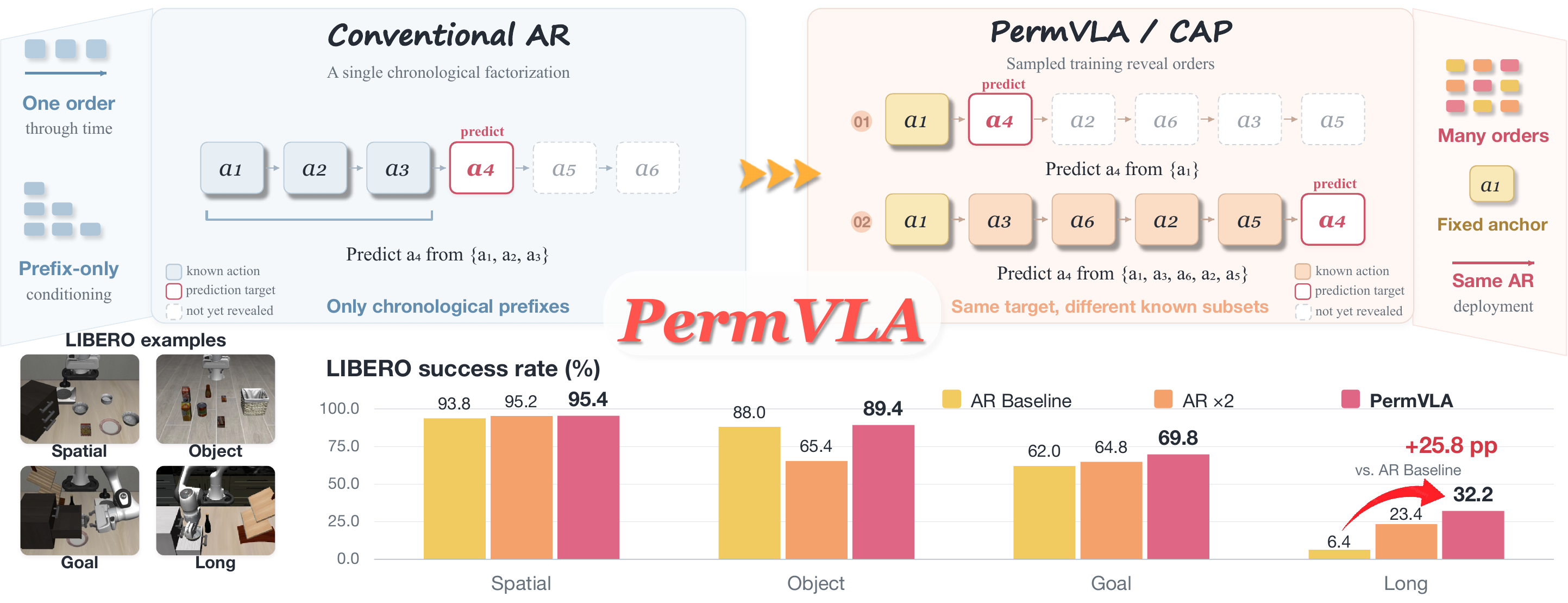}
  \caption{Overview of \method.  Autoregressive VLA training conventionally
  factorizes each expert action chunk in chronological order, so every
  action slot is always predicted from the same teacher-forced prefix.  \method{} treats
  this reveal order as a training-only degree of freedom: the same physical
  actions are supervised under multiple valid conditioning subsets, while
  the expert target, action identities, and chronological autoregressive
  deployment remain unchanged.}
  \label{fig:teaser}
\end{figure*}

We instantiate this idea as \method, a deliberately minimal VLA that
conditions a diagonal-Gaussian autoregressive action head on pretrained
vision-language model (VLM) embeddings.  This design prioritizes controlled
analysis and attribution of causally anchored permutation (CAP)
regularization; closed-loop success measures its contribution within the
fixed architecture.  CAP adds an auxiliary permutation likelihood to the
standard LTR objective.  The tunable
prefix length \(K\) fixes the first \(K\) actions in physical-time order;
\(K=0\) allows every reveal order.  The shared policy minimizes
\((1-\lambda)\mathcal L_{\mathrm{LTR}}+\lambda\mathcal L_{\mathrm{CAP}}\),
then deploys through deterministic LTR receding-horizon control.
Each expert chunk supplies prediction problems conditioned on different known
action subsets, with physical action identities and the deployment procedure
unchanged.  We call this \emph{conditional-set augmentation}.
Figure~\ref{fig:teaser} illustrates these conditioning patterns and
summarizes their effect in matched LIBERO comparisons.

Permutation modeling has established foundations in orderless density
estimation, XLNet, and autoregressive diffusion
\citep{uria2014deep,yang2019xlnet,hoogeboom2022ardm}.  Random autoregressive
factorization has also been applied to same-time multi-agent trajectories
\citep{rudolph2022conditional}, while Uni[MASK], MaskDP, and Masked Trajectory
Models learn completion under variable masks
\citep{carroll2022unimask,liu2022maskdp,wu2023mtm}.  We assign alternative
factorizations a distinct VLA role: a training-only regularizer over future
physical actions, with a tunable chronological prefix and paired with
the exact LTR deployment objective.  This formulation makes factorization
order directly measurable under a fixed controller.

Sample-matched LTR-1pass and decoder-compute-matched LTR-2pass controls isolate
the training objective.  LIBERO~\citep{liu2023libero} tests in-distribution
control, LIBERO-Plus~\citep{fei2026liberoplus} tests zero-shot
out-of-distribution (OOD) robustness, and CALVIN~\citep{mees2022calvin} tests
whether the gains extend to another benchmark.

In the matched reference experiments, CAP improves closed-loop success over
both LTR controls across these benchmarks.  Likelihood diagnostics show
greater agreement across reveal orders while retaining the fit under LTR,
connecting the training objective to a measurable change in the learned policy.

This approach suggests a broader design principle: varying known action
subsets can construct auxiliary objectives for other action generators while
retaining their deployment procedures.  We illustrate it with
\textsc{PermDiffusion}, which conditions on clean action subsets and denoises
the remaining slots.

Our contributions are:
\begin{itemize}
  \item We identify action-chunk factorization order as a previously
  overlooked VLA inductive-bias axis and formulate its regularization effect
  as conditional-set augmentation.
  \item We introduce CAP, an auxiliary factorization objective with tunable
  prefix length that diversifies action conditioning during training and
  preserves standard LTR deployment.
  \item Matched controls, parameter studies, OOD evaluation, and CALVIN
  demonstrate the empirical benefits.  Likelihood diagnostics provide evidence
  of agreement across reveal orders, and a diffusion extension illustrates
  the broader design space.
\end{itemize}

%% file: sections/02_related_work.tex
\section{Related Work}
\label{sec:related}

\paragraph{VLA action modeling.}
RT-1 and RT-2 established scalable vision-language robot policies with
tokenized control \citep{brohan2022rt1,brohan2023rt2}; Octo and OpenVLA
extended open robot pretraining across datasets and embodiments
\citep{octo2024,kim2024openvla}.  Recent action generators include the Action
Chunking with Transformers (ACT) policy \citep{zhao2023act}, conditional denoising in Diffusion
Policy \citep{chi2023diffusion}, flow matching in \(\pi_0\)
\citep{black2024pi0}, and compressed action tokens in FAST
\citep{pertsch2025fast}.  OpenVLA-OFT, WorldVLA, \(\pi_{0.5}\), and SmolVLA
further study chunk decoding, auxiliary prediction, heterogeneous
pretraining, and asynchronous execution
\citep{kim2025oft,cen2025worldvla,pi2025pi05,shukor2025smolvla}.  CAP fixes the
backbone and continuous action generator to isolate factorization order as a
training intervention.

\paragraph{Order-agnostic and masked sequence modeling.}
Orderless Neural Autoregressive Distribution Estimation (NADE) trains one
density estimator across autoregressive orders \citep{uria2014deep}; XLNet
combines permutation language modeling with two-stream attention
\citep{yang2019xlnet}; and autoregressive diffusion models (ARDMs) connect
order-agnostic autoregression to absorbing diffusion
\citep{hoogeboom2022ardm}.  \citet{rudolph2022conditional} carry the same
mechanism into trajectory modeling, pairing random autoregressive
factorizations with conditional two-stream attention for the same-time
movements of multiple agents.  Uni[MASK], MaskDP, and Masked Trajectory Models
learn behavior from variable trajectory masks
\citep{carroll2022unimask,liu2022maskdp,wu2023mtm}.  In each case, permutation
or masking constitutes the primary training objective itself, defined over
unordered variable sets or completion-style tasks.

\paragraph{Permutation training as a VLA regularizer.}
Our contribution is to import the permutation idea of XLNet into VLA learning
in the form of a regularizer: the permuted likelihood enters as an auxiliary
term on top of the chronological density objective.  Two properties of
physical-world action imitation make this transfer practical.  The first is
\emph{computational feasibility}: expert action chunks are short
(\(H=8\)--\(16\) slots), so evaluating normalized likelihoods under \(M\)
sampled reveal orders adds modest decoder computation, while the parameter
count and deployment cost remain fixed.  The second is \emph{temporal
adherence}: the chronological LTR objective remains the dominant training
path and the only deployment path, so the policy keeps physical-time ordering
as its primary interface to control.  The auxiliary CAP term then predicts
each target action from varied known subsets, discouraging over-reliance on
one generation path and on the visual correspondences specific to that path.
CAP assigns an ordered, normalized likelihood to each sampled order within one
future action chunk, whereas masked-completion objectives score unordered
variable sets with completion-style losses.
The tunable prefix length \(K\) controls the order family, including fully
unrestricted orders at \(K=0\).  Our reference configuration draws only
\(M=1\) sampled order per example, with a \(M=4\) ablation, so the
computational-efficiency argument above is validated only in the low-\(M\)
regime; scaling \(M\) further and mapping the resulting
computation--regularization trade-off remains an open direction. 

%% file: sections/03_method.tex
\section{Method}
\label{sec:method}

\subsection{Problem Formulation}

At control time \(t\), the policy receives
\(\context_t=(\ell,o_{\leq t},s_{\leq t})\), consisting of a language
instruction, the configured visual history, and robot proprioception.  It
predicts a length-\(H\) action chunk
\(\actions_t=(a_t,\ldots,a_{t+H-1})\), with
\(a_i\in\mathbb{R}^{D}\), by learning
\(p_\theta(\actions_t\mid\context_t)\).
Every conditioning variable is available through the control interface or an
action value revealed earlier in the selected factorization.

Let \(z=(z_1,\ldots,z_H)\) be a permutation of the physical action-slot indices
\(\{1,\ldots,H\}\).
Write
\(\mathcal R_{z,<j}=\{(z_m,a_{t+z_m-1})\}_{m<j}\) for the indexed actions
revealed before factorization step \(j\).
The corresponding autoregressive factorization is
\begin{equation}
  p_\theta(\actions_t\mid\context_t,z)
  =
  \prod_{j=1}^{H}
  p_\theta\!\left(
    a_{t+z_j-1}\mid\context_t,\mathcal R_{z,<j}
  \right).
  \label{eq:general-factorization}
\end{equation}
The physical index \(z_j\) is attached to every action value through a learned
time embedding.
The permutation changes reveal order and preserves physical execution time.

\subsection{Permutation Modeling with a Tunable Anchor}

For \(K\in\{0,\ldots,H\}\), we define the family of \(K\)-anchored orders
\begin{equation}
  \Pi_K
  =
  \begin{cases}
    \mathfrak S_H, & K=0, \\[2pt]
    \left\{z\in\mathfrak S_H:\ z_j=j,\quad
    \forall j\in\{1,\ldots,K\}\right\}, & 1\leq K\leq H,
  \end{cases}
  \label{eq:anchored-family}
\end{equation}
where \(\mathfrak S_H\) is the set of all permutations of the \(H\) action
slots.  When \(K=0\), the anchored prefix is empty and all \(H!\) reveal
orders are admissible.  For \(K>0\), the first \(K\) reveal positions are
fixed to physical slots \(1,\ldots,K\), while the remaining slots can appear
in any order.  Thus \(K=1\) fixes the first action and \(K=H\) recovers LTR.
Increasing \(K\) narrows the family by fixing a longer chronological prefix,
controlling the balance between reveal-order
diversity and temporal structure in the auxiliary loss.  Every setting retains
the chronological LTR objective and deployment path.

\paragraph{Conditional-set augmentation.}
\label{sec:cap-mechanism}

For a physical suffix slot \(i>K\), let
\(S_z(i)=\{z_m:m<r_z(i),\,z_m>K\}\) collect the suffix slots revealed before
\(i\) under order \(z\).  CAP predicts \(a_{t+i-1}\) from \(\context_t\), the
prefix \(a_{t:t+K-1}\) (empty when \(K=0\)), and the revealed actions indexed by
\(S_z(i)\), where \(r_z(i)\) is the rank of slot \(i\) in \(z\).
The target retains its physical index while the conditioning set
changes, turning one expert chunk into sampled missing-action completion
problems.

LTR and CAP fit the same expert action distribution through different
factorizations.  LTR predicts a slot from its chronological prefix; CAP also
predicts it from sampled revealed subsets.  Sharing decoder parameters couples
these prediction problems.  We hypothesize that this training pressure favors
dependencies useful across conditioning sets, reducing reliance on one
generation path.

For each target action, the two-stream attention mask exposes only action
values placed earlier in the sampled reveal order, preventing target leakage
for every \(K\).  At deployment, the controller generates and executes actions
chronologically using the observed context and its own predicted prefix.

\subsection{Reference Architecture}

\begin{figure*}[t]
  \centering
  \includegraphics[width=1\linewidth]{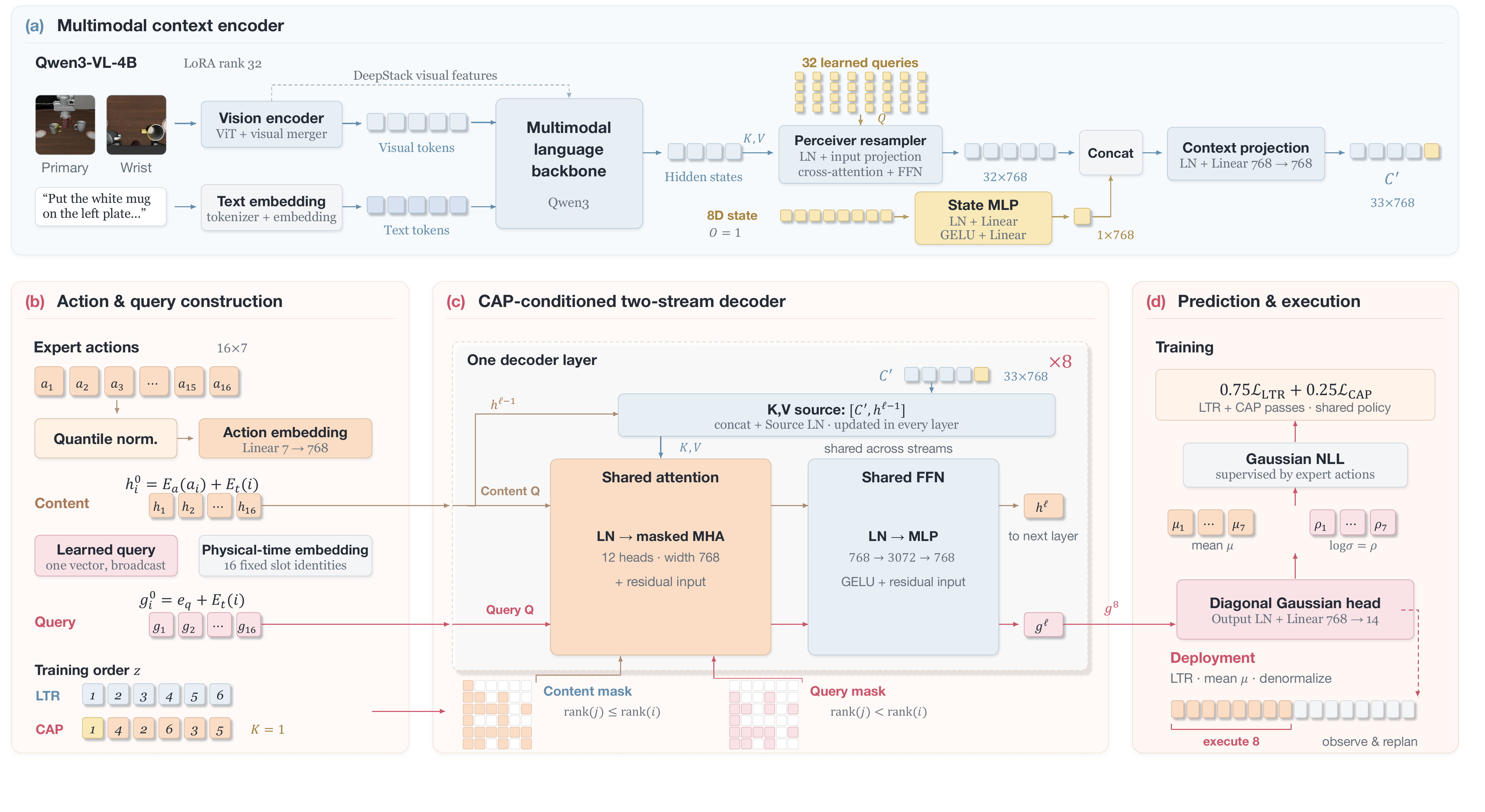}
  \caption{
    \method architecture.  A pretrained VLM with a Perceiver-style resampler
    produces context tokens that condition a two-stream action decoder; the
    sampled reveal order enters only through the attention masks during
    training, while deployment always executes a chronological action prefix.
  }
  \label{fig:architecture}
\end{figure*}

Figure~\ref{fig:architecture} shows the reference policy.  Qwen3-VL-4B
\citep{bai2025qwen3vl} with low-rank adaptation (LoRA) encodes the instruction
and two camera views.  A Perceiver-style resampler uses 32 learned query tokens
to cross-attend to the variable-length VLM output and produce a fixed-size
context sequence.  Projected proprioception tokens are appended to this
sequence.  An eight-layer, 67.0M-parameter conditional Transformer predicts a
diagonal Gaussian action
distribution from quantile-normalized expert actions.  CAP and both LTR
controls share this architecture and differ only in their training
objectives.  Appendix~\ref{app:complete-experimental-parameters} reports the
full dimensions and optimization settings.


\paragraph{Conditional two-stream attention.}
We implement each factorization with XLNet-style content and query streams
\citep{yang2019xlnet}.  Physical-time embeddings preserve action-slot identity,
while attention masks are constructed directly from the sampled reveal order.
A content state reads earlier revealed actions and its own action value; the
corresponding query state reads only earlier actions, with its target masked.
Both streams share Transformer parameters, and the Gaussian head reads the
query states.  Appendix~\ref{app:two-stream} gives the exact mask construction.

\subsection{Training Objective}
\label{sec:training-objective}

We now switch to chunk-local indices: \(a_i\equiv a_{t+i-1}\) denotes the
\(i\)-th physical action slot of the chunk \(\actions_t\).
For each physical slot \(i\), the Gaussian head predicts mean
\(\mu_i^{(z)}\) and log standard deviation \(\rho_i^{(z)}\).  We write its
slot-level negative log likelihood (NLL) as
\(\ell_i^{(z)}=-\log\mathcal N(a_i;\mu_i^{(z)},
\operatorname{diag}(e^{2\rho_i^{(z)}}))\).  Let
\(z^{\mathrm{LTR}}=(1,\ldots,H)\).  The standard LTR objective used at
deployment and the Monte Carlo CAP objective are
\begin{align}
  \ltrloss
  &=
  \frac{1}{BH}
  \sum_{b=1}^{B}\sum_{i=1}^{H}
  \ell_{b,i}^{(z^{\mathrm{LTR}})}, 
  \label{eq:ltr-loss}
  \\
  \caploss
  &=
  \frac{1}{MBH}
    \sum_{m=1}^{M}\sum_{b=1}^{B}\sum_{j=1}^{H}
    \ell_{b,z_{b,m,j}}^{(z_{b,m})}.
  \label{eq:cap-loss}
\end{align}
Each example draws \(M\) independent orders \(z_{b,m}\sim\Pi_K\).
Increasing \(M\) exposes a chunk to more conditioning subsets and reduces the
Monte Carlo variance of the CAP estimate.  It increases training-time action
decoder work; the parameter count and LTR inference cost remain fixed.
Our final training objective is
\begin{equation}
  \mathcal{L}_{\mathrm{PermVLA}}
  =
  (1-\lambda)\ltrloss
  +
  \lambda\caploss,
  \label{eq:total-loss}
\end{equation}
with \(K=1\) and \(\lambda=0.25\) in the reference configuration.  The value
of \(M\) is one in this configuration; additional sampled orders isolate its effect in
the targeted configurations reported in
Appendix~\ref{app:targeted-libero-diagnostics}.
The LTR and CAP terms reuse one multimodal context encoding and share every
action-decoder parameter.  Each sample contributes one chronological LTR
prediction problem and \(M\) missing-action completion problems constructed
from sampled known-action subsets.  The LTR term directly supervises
deployment; the CAP term trains the same policy representation across those
different subsets.

The exact chain-rule identity, its finite-capacity interpretation, and
testable regularization hypotheses are developed in
Appendix~\ref{app:cap-theory}.  That analysis separates what follows exactly
from likelihood factorization from the empirical hypothesis that shared
parameters trained across known-action subsets rely less on a single reveal
order.

%% file: sections/04_experiments.tex
\section{Experiments}
\label{sec:experiments}

The experiments proceed in four stages.  We first establish
the closed-loop capability of the compact VLM with a Gaussian autoregressive
policy and isolate
CAP from data exposure and decoder computation.  We then test whether the
effect persists under LIBERO-Plus distribution shifts and on CALVIN, before
measuring how its joint likelihood changes across reveal orders.

\subsection{Experimental Setup}
\label{sec:experimental-setup}

\paragraph{Reference policy and protocol.}
The fixed policy combines Qwen3-VL-4B-Instruct with LoRA, a Perceiver-style
resampler, and a 67M-parameter diagonal-Gaussian autoregressive action head.
It conditions on language, primary and wrist RGB, and proprioception, and is
trained in one supervised adaptation stage.  Appendix
\ref{app:complete-experimental-parameters} specifies the complete data, model,
optimization, hardware, and evaluation protocol.

\paragraph{Matched controls.}
Every causal comparison holds the initialization, architecture, data,
optimizer, schedule, normalization, model selection, and deployment fixed.
LTR-1pass uses one LTR objective; LTR-2pass adds a weighted second LTR pass to
match decoder work and independently sampled dropout; \method replaces that
pass with \(M\) CAP reveal orders.  All three deploy through deterministic
chronological LTR generation.

\paragraph{Evaluation.}
LIBERO trains a separate policy for each suite and evaluates 500 matched
episodes per suite.  LIBERO-Plus evaluates those policies zero-shot with one
rollout for each of 10,030 released perturbation instances, and
CALVIN evaluates 1,000 fixed five-instruction chains under ABC$\rightarrow$D.
Published systems provide performance context; the matched controls identify
the effect of CAP and test its persistence across distribution shift and
benchmark.

\subsection{LIBERO Evaluation of the Reference Policy}
\label{sec:competitive-simple-policy}

Table~\ref{tab:libero-context} places the compact policy among representative
published systems under their original protocols.  Matched attribution follows
in Table~\ref{tab:cap-controlled-sweep}.

\begin{table*}[!t]
  \centering
  \caption{LIBERO success rate (\%).  Published rows retain their original
  protocols.  Diffusion Policy and Octo values follow \citet{kim2024openvla},
  \(\pi_0\) follows \citet{driess2025knowledge}, and the remaining values
  follow their cited sources.  Our rows use suite-specific policies trained
  for 30k updates and evaluated over 500 episodes per suite.  Labels abbreviate
  configuration values; e.g., M1/H16/O1/E8 denotes
  \(M=1,H=16,O=1,e=8\).}
  \label{tab:libero-context}
  \tablebodyfont
  \renewcommand{\arraystretch}{0.96}
  \resizebox{\textwidth}{!}{%
  \begin{tabular}{lccccccc}
    \toprule
    Model & Robot pretraining & Action generator & Spatial & Object & Goal & Long & Mean \\
    \midrule
    \multicolumn{8}{l}{\textit{Representative published policies}} \\
    Diffusion Policy~\citep{chi2023diffusion} & None & Diffusion & 78.3 & 92.5 & 68.3 & 50.5 & 72.4 \\
    Octo~\citep{octo2024} & Yes & Diffusion & 78.9 & 85.7 & 84.6 & 51.1 & 75.1 \\
    OpenVLA~\citep{kim2024openvla} & Yes & Token autoregression & 84.7 & 88.4 & 79.2 & 53.7 & 76.5 \\
    $\pi_0$~\citep{black2024pi0} & Yes & Flow matching & 96.8 & 98.8 & 95.8 & 85.2 & 94.2 \\
    SmolVLA-0.45B~\citep{shukor2025smolvla} & None & Flow matching & 90.0 & 96.0 & 92.0 & 71.0 & 87.3 \\
    WorldVLA-256~\citep{cen2025worldvla} & None & Token autoregression + world model & 85.6 & 89.0 & 82.6 & 59.0 & 79.1 \\
    RIPT-VLA~\citep{tan2025ript} & Yes & Regression + online RL & 98.6 & 98.6 & 99.0 & 93.8 & 97.5 \\
    \midrule
    \multicolumn{8}{l}{\textit{Our controlled study}} \\
    \method, M1/H8/O1 & None & 67M Gaussian autoregression & 88.6 & 97.0 & 90.2 & 34.2 & 77.5 \\
    \method, M1/H16/O1/E8 & None & 67M Gaussian autoregression & 95.4 & 89.4 & 69.8 & 32.2 & 71.7 \\
    \method, M1/H16/O1/E16 & None & 67M Gaussian autoregression & 91.8 & 82.8 & 77.6 & 31.6 & 71.0 \\
    \bottomrule
  \end{tabular}%
  }
\end{table*}

The compact Gaussian policy reaches the range of established VLA systems, so
its capacity is sufficient for controlled comparisons among training
objectives.  Shorter horizons give the best overall balance, while temporal
settings redistribute suite performance and leave long-horizon control as the main gap.

\subsection{Controlled Evaluation of the CAP Regularizer}
\label{sec:controlled-sweep}

Table~\ref{tab:cap-controlled-sweep} uses
\(M=1,H=16,O=1,e=8,K=1,\lambda=0.25\) as its reference: one temporal
observation from each camera, a 16-step prediction horizon, and eight executed
actions.  Objective rows share the entire protocol; other rows change only the
named setting, with execution variants reusing the trained policy.  Appendix
\ref{app:targeted-libero-diagnostics} reports targeted diagnostics and completed
configurations.

\begin{table*}[!t]
  \centering
  \caption{Objective and temporal comparisons on LIBERO success rate
  (\%).  The reference is
  \(M=1,H=16,O=1,e=8,K=1,\lambda=0.25\), 30k updates, seed zero, and 500
  episodes per suite.  Every non-objective row changes only the named value.}
  \label{tab:cap-controlled-sweep}
  \tablebodyfont
  \renewcommand{\arraystretch}{0.96}
  \setlength{\tabcolsep}{6pt}
  \begin{tabular}{llccccc}
    \toprule
    Controlled factor & Setting & Spatial & Object & Goal & Long & Mean \\
    \midrule
    Objective & LTR-1pass & 93.8 & 88.0 & 62.0 & 6.4 & 62.6 \\
              & LTR-2pass & 95.2 & 65.4 & 64.8 & 23.4 & 62.2 \\
              & \method reference & 95.4 & 89.4 & 69.8 & 32.2 & 71.7 \\
    \midrule
    CAP orders $M$ & 4 & 93.6 & 87.4 & 66.0 & 42.0 & 72.3 \\
    \midrule
    Horizon $H$ & 8 & 88.6 & 97.0 & 90.2 & 34.2 & 77.5 \\
    \midrule
    History $O$ & 2 & 85.0 & 96.0 & 75.4 & 38.4 & 73.7 \\
    \midrule
    Execute $e$ & 16 (\method) & 91.8 & 82.8 & 77.6 & 31.6 & 71.0 \\
                & 16 (LTR-1pass) & 90.8 & 62.4 & 63.8 & 8.2 & 56.3 \\
    \midrule
    Anchor $K$ & 0 & 95.6 & 76.6 & 79.6 & 35.8 & 71.9 \\
               & 8 & 93.2 & 79.0 & 65.8 & 36.0 & 68.5 \\
    \midrule
    CAP weight $\lambda$ & 0.50 & 93.0 & 85.6 & 77.4 & 40.6 & 74.2 \\
    \bottomrule
  \end{tabular}
\end{table*}

The \(K=1\) reference exceeds both LTR controls on every suite; a second LTR
pass leaves mean success at the one-pass level.  This separates reveal-order
diversity from decoder computation and dropout exposure.  The largest gap on Long supports the view
that varied conditioning subsets improve dependencies among future actions
needed for extended closed-loop behavior.

Temporal settings redistribute success across task families.  A second
observation helps Object, Goal, and Long while reducing Spatial, and the
reference CAP--LTR ordering persists at both execution lengths.
With \(K=0\), the four-suite mean stays close to \(K=1\): Goal and Long
improve, Object declines, and Spatial changes little.  \(K=8\) also improves
Long while lowering the other suites.  These outcomes identify prefix length
as a task-dependent regularization parameter, with both anchored and
unrestricted orders supporting effective training.
Increasing \(M\) favors Long; a larger CAP weight improves Goal and Long.
These interventions suggest that the strength and diversity of the auxiliary
constraints affect composition-heavy behavior in distinct ways.
Appendix Tables~\ref{tab:libero-suite-deltas} and
\ref{tab:libero-completed-configs} summarize the suite effects;
Tables~\ref{tab:table2-task-spatial}--\ref{tab:table2-task-long} localize
them by task.

The objective and temporal rows answer different questions.  The matched
objective block provides the causal comparison for CAP; the \(H\), \(O\), and
\(e\) rows characterize how that gain interacts with control timescale and
available history.  Their suite-dependent optima show that the CAP advantage
holds across temporal configurations, and they expose a second design
axis that can be optimized after selecting the training objective.

\paragraph{Offline fit and closed-loop success diverge.}
On Long, LTR fits expert actions slightly more closely while CAP produces much
stronger closed-loop control.  This reversal suggests that training from
multiple known-action subsets improves coherent continuation or recovery once
the policy visits its own states.
Section~\ref{sec:factorization-diagnostic} tests whether CAP reduces dependence
on reveal order; perturb-and-recover evaluation can distinguish continuation
from recovery.  Appendix Table~\ref{tab:ltr-cap-long-diagnostics} reports the
training diagnostics.

\subsection{Zero-Shot Robustness on LIBERO-Plus}
\label{sec:liberoplus}

LIBERO-Plus evaluates the standard-LIBERO policies through direct zero-shot
transfer.  Table~\ref{tab:liberoplus-main} reports success by perturbation category and
includes representative published VLAs for context.  Our controlled rows use
the same model-selection rule and matched treatments as the controlled LIBERO
study.

\begin{table*}[!t]
  \centering
  \caption{Zero-shot LIBERO-Plus success rate (\%).  ``Original'' is the
  corresponding standard-LIBERO score; the remaining columns partition the 10,030
  perturbation instances.  Published rows are transcribed from Table~2 of
  \citet{fei2026liberoplus} under their published protocols.  We retain
  representative token, regression, and fast-action baselines here.  Our rows
  use standard-LIBERO training trajectories and frozen policies.}
  \label{tab:liberoplus-main}
  \tablebodyfont
  \resizebox{\textwidth}{!}{%
  \begin{tabular}{lcccccccc}
    \toprule
    Method & Original & Camera & Robot & Language & Light & Background & Noise & Layout \\
    \midrule
    \multicolumn{9}{l}{\textit{Published contextual references}} \\
    OpenVLA        & 76.5 &  0.8 &  3.5 & 23.0 &  8.1 & 34.8 & 15.2 & 28.5 \\
    $\pi_0$-FAST    & 85.5 & 65.1 & 21.6 & 61.0 & 73.2 & 73.2 & 74.4 & 68.8 \\
    WorldVLA        & 79.1 &  0.1 & 27.9 & 41.6 & 43.7 & 17.1 & 10.9 & 38.0 \\
    RIPT-VLA        & 97.5 & 55.2 & 31.2 & 77.6 & 88.4 & 91.6 & 73.5 & 74.2 \\
    \midrule
    \multicolumn{9}{l}{\textit{Our controlled study}} \\
    LTR-1pass & 62.6 & 22.2 & 29.5 & 42.2 & 56.8 & 45.5 & 20.3 & 47.7 \\
    LTR-2pass & 62.2 & 23.1 & 25.9 & 50.0 & 42.8 & 45.1 & 23.5 & 42.0 \\
    \method M1/H8/O1 & 77.5 & 23.3 & 34.3 & 56.2 & 63.9 & 50.0 & 21.2 & 50.6 \\
    \method M1/H16/O1 & 71.7 & 26.7 & 31.4 & 62.7 & 61.2 & 50.6 & 26.0 & 50.0 \\
    \bottomrule
  \end{tabular}%
  }
\end{table*}

CAP improves all seven perturbation aggregates against both matched LTR
controls and raises the zero-shot mean in every suite.  It wins 20 of 28
suite--perturbation combinations against LTR-1pass and 25 of 28 against compute-matched
LTR-2pass.  Since these shifts are absent from training, the gain transfers
beyond the training trajectories: the same policy remains more successful when
visual observations, language instructions, or robot appearance changes.


The gains are broadest for Language and Background, while Long improves under
every perturbation family. This pattern links CAP to more stable action
continuation under visual and semantic change. Failure-stage annotations can
further separate perception from recovery errors. The compact \(H=8\) policy
also exceeds OpenVLA and WorldVLA across every shift family; RIPT-VLA shows
the additional contribution of large-scale pretraining and
post-training.  Appendix Tables~\ref{tab:liberoplus-suite}--%
\ref{tab:liberoplus-ltr2-suite} report the suite-resolved matrices.

\subsection{Cross-Benchmark Transfer on CALVIN}
\label{sec:calvin}

CALVIN changes the visual domain, demonstrations, control horizon, and
evaluation unit to five-instruction chains.  Table~\ref{tab:calvin-main} tests
whether the matched LIBERO ordering transfers; Diffusion Policy-CNN and OpenVLA
provide external context.

\begin{table*}[!t]
  \centering
  \caption{CALVIN ABC$\rightarrow$D evaluation on 1,000 fixed
  five-instruction chains.  Success@\(k\) is the percentage completing at
  least \(k\) consecutive instructions.  External rows use independently
  published training and model-selection protocols and provide context; bold
  denotes the best result within each block.}
  \label{tab:calvin-main}
  \tablebodyfont
  \resizebox{\textwidth}{!}{%
  \begin{tabular}{lcccccc}
    \toprule
    Method & Avg. length & Success@1 & Success@2 & Success@3 & Success@4 & Success@5 \\
    \midrule
    \multicolumn{7}{l}{\textit{External protocol context}} \\
    Diffusion Policy-CNN~\citep{reuss2025mode} & 1.350 & 63.5 & 35.3 & 19.4 & 10.7 & 6.4 \\
    OpenVLA~\citep{kim2024openvla,xiao2026avavla} & \textbf{3.270} & \textbf{91.3} & \textbf{77.8} & \textbf{62.0} & \textbf{52.1} & \textbf{43.5} \\
    \midrule
    \multicolumn{7}{l}{\textit{Matched Gaussian autoregressive controls}} \\
    LTR-1pass & 1.090 & 61.0 & 27.6 & 12.2 & 5.9 & 2.3 \\
    LTR-2pass & 0.694 & 40.3 & 18.4 & 7.6 & 2.3 & 0.8 \\
    \method & \textbf{1.364} & \textbf{63.7} & \textbf{36.8} & \textbf{19.7} & \textbf{10.1} & \textbf{6.1} \\
    \bottomrule
  \end{tabular}
  }
\end{table*}

CAP leads both matched controls at every chain depth and reproduces the LIBERO
ordering, CAP $>$ LTR-1pass $>$ LTR-2pass.  Its relative advantage grows with
chain depth: the absolute gap is largest at Success@2, while the ratio of CAP
to LTR-1pass success rises from 1.04 at Success@1 to 2.65 at Success@5,
indicating a cumulative benefit in long closed-loop sequences and
transfer across demonstration data, visual observations, and task sequences.

LTR-2pass has the lowest logged pointwise action error and the weakest control,
repeating the LIBERO Long reversal.  The two benchmarks therefore separate CAP
from one-step fitting and extra decoder computation, and associate
reveal-order diversity with dependencies that remain useful as errors compound.
CAP also matches Diffusion Policy-CNN in average completed chain length and
in the pattern across Success@\(1\) through Success@\(5\).  OpenVLA remains
stronger, illustrating the additional capability
from large-scale robot pretraining; the matched rows isolate the objective's
contribution.

Because Success@\(k\) is nested, the widening separation means that CAP retains
a larger fraction of successful trajectories as instruction depth increases.
This mirrors the Long result and locates the benefit after the first successful
interaction, where the policy must continue from states produced by its own
actions.  The benefit therefore extends beyond completing the first instruction.

\subsection{Factorization Sensitivity Diagnostic}
\label{sec:factorization-diagnostic}

CAP predicts more consistent joint likelihoods for the same expert chunk
across valid reveal orders.  We measure this factorization sensitivity as
\begin{equation}
  \mathcal S(\theta)
  =
  \mathbb E_{(\context,\actions)}
  \mathbb E_{z,z'\sim\Pi_K}
  \left[
    \left(
      \log q_\theta^z(\actions\mid\context)
      -
      \log q_\theta^{z'}(\actions\mid\context)
    \right)^2
  \right],
  \qquad
  \mathcal S_H(\theta)=\frac{\mathcal S(\theta)}{H^2}.
  \label{eq:factorization-sensitivity}
\end{equation}
where \(q_\theta^z\) is the normalized joint density under order \(z\);
\(\mathcal S_H\) measures disagreement between average per-step log
likelihoods and is zero under reveal-order agreement.  We compute it after
training on identical task-balanced chunks and orders.

CAP lowers sensitivity by several orders of magnitude across all suites while
preserving chronological LTR likelihood; LTR-2pass remains sensitive.  Higher
\(\lambda\) jointly lowers sensitivity and raises Goal and Long success, and
cross-evaluation shows that agreement follows the anchor family seen in
training.  In contrast, \(M=4\) improves Long without further reducing
\(\mathcal S_H\), confirming that the diagnostic isolates reveal-order
agreement specifically; other determinants of control can move independently.
Two caveats further bound
its interpretation: conditional independence among the action slots would
minimize \(\mathcal S_H\) trivially without implying coordinated execution,
and the \(K=8\) intervention raises \(\mathcal S_H\) under \(\Pi_1\) while
improving Long success.  We therefore use the diagnostic to
verify that the policy has internalized agreement over its trained permutation
family, and rely on closed-loop success to rank control performance.  Appendix
Tables~\ref{tab:factorization-sensitivity}--%
\ref{tab:anchor-family-cross-eval} report the protocol and interventions.

%% file: sections/05_discussion.tex
\section{Discussion}
\label{sec:discussion}

\paragraph{A neglected regularization axis.}
CAP exposes factorization order as a controllable choice in VLA supervision.
The factorization-sensitivity diagnostic shows that CAP makes a chunk's joint
likelihood less dependent on its reveal order.  LIBERO-Plus and CALVIN show
that the resulting conditional dependencies among future actions remain useful
outside the training environments.  Together, these results motivate studying
alternative chain-rule factorizations and routinely evaluating OOD robustness
and recovery.  As an extension, future work can study variations of CAP such
as order-free random-mask completion, soft or continuous relaxations of the
binary order masks, consistency penalties that directly shrink disagreement
between orders, and finer-grained reveal orders over individual action
dimensions.  Comparing
these variants under the same matched controls would isolate which ingredient
of conditional-set augmentation drives the observed gains.

\paragraph{Temporal scope and adaptive factorization.}
Closed-loop control also depends on partial observability, policy-induced
states, recovery, and executed prefix \(e\).  Suite-dependent effects motivate
learned reveal-order distributions, selecting \(K\), including zero, from
predictive uncertainty, and curricula that move from LTR to more permuted
orders.  Rollouts without teacher forcing and perturb-and-recover tests can
separate continuation, likelihood calibration, and recovery.

\paragraph{Action generators and scaling.}
The Gaussian autoregressive policy enables controlled attribution, while
\textsc{PermDiffusion} applies the same idea by treating sampled action slots
as clean inputs when denoising the others.  Matched diffusion, flow,
token-autoregressive, multi-effector, and pretrained-VLA studies can test
scaling, regularization strength, and which network modules should receive CAP
gradients.

\paragraph{Limitations.}
Our controlled comparisons center on Gaussian autoregressive policies in
simulation; real-robot transfer and interaction with large-scale robot
pretraining remain open.  Gains vary across tasks and temporal settings,
motivating task-aware regularization.  Factorization sensitivity measures
agreement over sampled reveal orders; identifying which learned dependencies
drive OOD gains requires targeted interventions.  \textsc{PermDiffusion}
provides a feasibility example, and matched diffusion controls are needed to
quantify the auxiliary objective's benefit.

%% file: sections/06_conclusion.tex
\section{Conclusion}
\label{sec:conclusion}

This work identifies action-chunk factorization order as a practical VLA
regularization axis.  CAP trains over sampled reveal orders with a tunable
prefix and deploys chronologically.  Matched controls isolate its gains from computation
and pointwise fit; LIBERO-Plus and CALVIN show consistent generalization across
environments and datasets.  The diagnostic verifies that the trained policy
agrees across reveal orders.  These findings show that changing how a fixed
demonstration is factorized can improve the learned policy without altering
deployment-time inference.  They position factorization choice as a general
tool for designing VLA objectives, complementary to model scaling and robot
pretraining.  Future work should learn task-adaptive reveal orders, transfer
CAP to diffusion and flow policies, and test its behavior in larger pretrained
VLAs, broader benchmarks, and multi-effector control.

%% file: sections/appendix_a_experimental_details.tex
\section{Experimental Details and Suite-Resolved Results}
\label{app:experimental-details}

The main paper isolates the CAP mechanism with matched LTR controls and
controlled temporal comparisons.  This appendix collects the four-suite
LIBERO configurations, the planned backbone-replacement control,
suite-resolved LIBERO-Plus breakdowns, training diagnostics, parameter
semantics, and the consolidated reproducibility settings behind the
main-paper conclusions.  Appendix~\ref{app:robot-visualization} complements
these aggregate results with full robot rollouts and an audit of
demonstration endings and success criteria.
Compact configuration labels concatenate parameter values; for example,
M1/H16/O1/E8 denotes \(M=1,H=16,O=1,e=8\).

\subsection{Backbone Replacement Ablation}
\label{app:backbone}

This ablation keeps the action decoder, CAP objective, data, optimization
budget, and evaluation protocol fixed while replacing the multimodal
backbone.  It tests whether the same factorization regularizer and Gaussian
action policy remain effective with a different visual-language
representation.

\begin{table}[!ht]
  \centering
  \caption{Backbone replacement under the fixed CAP
  \(M=1,H=16,O=1,e=8\) protocol.  Entries are 500-episode LIBERO success
  rates (\%).}
  \label{tab:backbone}
  \tablebodyfont
  \setlength{\tabcolsep}{4pt}
  \begin{tabular}{lcccccc}
    \toprule
    Backbone & Adaptation & Spatial & Object & Goal & Long & Mean \\
    \midrule
    Qwen3-VL-4B-Instruct & LoRA-32 & 95.4 & 89.4 & 69.8 & 32.2 & 71.7 \\
    PaliGemma~2--3B & LoRA-32 & 90.0 & 82.8 & 79.0 & 23.4 & 68.8 \\
    \bottomrule
  \end{tabular}
\end{table}

PaliGemma~2--3B \citep{steiner2024paligemma2} provides a comparably scaled
representation family.  With the policy and training protocol fixed, it
improves Goal by 9.2 points while retaining strong Spatial and Object
performance.  This pattern shows that the CAP training construction remains
viable after replacing the VLM representation.  Its four-suite mean remains
within 2.9 points of the Qwen3-VL policy despite a lower Long score.

\subsection{Four-Suite LIBERO Configurations}
\label{app:libero-completed-configs}

Table~\ref{tab:libero-completed-configs} provides the absolute scores behind
the comparisons in
Table~\ref{tab:libero-suite-deltas}.  Targeted experiments that answer a
single deployment or mechanism question appear separately in
Tables~\ref{tab:long-execution-ablation}--\ref{tab:cap-gradient-routing}.

\begin{table*}[!t]
  \centering
  \caption{Four-suite LIBERO configurations (\%).  Every row uses
  30k updates and 500 evaluation episodes per suite.  The \(O=1\) block
  supports the main controlled analysis; the \(O=2\) block reports the
  observation-history comparison and the earlier \(M=4\) variants.}
  \label{tab:libero-completed-configs}
  \tablebodyfont
  \setlength{\tabcolsep}{4pt}
  \begin{tabular}{lcccccccccc}
    \toprule
    Objective & $M$ & $H$ & $O$ & $e$ & $K$ & Spatial & Object & Goal & Long & Mean \\
    \midrule
    \multicolumn{11}{l}{\textit{Main \(O=1\) configurations}} \\
    \method   & 1 & 8  & 1 & 8  & 1 & 88.6 & 97.0 & 90.2 & 34.2 & 77.5 \\
    LTR-1pass & 1 & 16 & 1 & 8  & 1 & 93.8 & 88.0 & 62.0 & 6.4  & 62.6 \\
    LTR-2pass & 1 & 16 & 1 & 8  & 1 & 95.2 & 65.4 & 64.8 & 23.4 & 62.2 \\
    \method\ ($\lambda=.25$) & 1 & 16 & 1 & 8 & 1 & 95.4 & 89.4 & 69.8 & 32.2 & 71.7 \\
    \method   & 4 & 16 & 1 & 8  & 1 & 93.6 & 87.4 & 66.0 & 42.0 & 72.3 \\
    \method\ ($\lambda=.50$) & 1 & 16 & 1 & 8 & 1 & 93.0 & 85.6 & 77.4 & 40.6 & 74.2 \\
    \method   & 1 & 16 & 1 & 8  & 0 & 95.6 & 76.6 & 79.6 & 35.8 & 71.9 \\
    \method   & 1 & 16 & 1 & 8  & 8 & 93.2 & 79.0 & 65.8 & 36.0 & 68.5 \\
    LTR-1pass & 1 & 16 & 1 & 16 & 1 & 90.8 & 62.4 & 63.8 & 8.2  & 56.3 \\
    \method   & 1 & 16 & 1 & 16 & 1 & 91.8 & 82.8 & 77.6 & 31.6 & 71.0 \\
    \midrule
    \multicolumn{11}{l}{\textit{\(O=2\) diagnostic configurations}} \\
    \method   & 1 & 16 & 2 & 8  & 1 & 85.0 & 96.0 & 75.4 & 38.4 & 73.7 \\
    \method   & 4 & 16 & 2 & 8  & 1 & 96.2 & 91.4 & 77.0 & 37.2 & 75.5 \\
    \method   & 4 & 16 & 2 & 16 & 1 & 85.6 & 83.0 & 77.6 & 31.8 & 69.5 \\
    \bottomrule
  \end{tabular}
\end{table*}

\subsection{Suite-Resolved Parameter Comparisons}
\label{app:suite-resolved-attribution}

Table~\ref{tab:libero-suite-deltas} reports signed changes for configurations
evaluated on all four suites.  The first panel contains direct one-factor
interventions, and the second records coupled
configurations for descriptive analysis.  Single-suite diagnostics are
isolated by their experimental question in the following subsection.

\begin{table*}[!t]
  \centering
  \caption{Signed change in LIBERO success rate (percentage points) for
  four-suite comparisons.  Each transition is treatment minus
  parent.  The one-factor panel changes only the parameter named in the first
  column; the coupled panel is descriptive.  Positive values denote
  improvement and negative values denote regression.  The Mean column averages
  the unrounded suite changes.}
  \label{tab:libero-suite-deltas}
  \tablebodyfont
  \renewcommand{\arraystretch}{0.96}
  \setlength{\tabcolsep}{4pt}
  \begin{tabular}{llccccc}
    \toprule
    Controlled change & Held fixed & Spatial & Object & Goal & Long & Mean \\
    \midrule
    \multicolumn{7}{l}{\textit{One-factor comparisons}} \\
    Objective: LTR-1pass $\rightarrow$ CAP & M1/H16/O1/E8 & +1.6 & +1.4 & +7.8 & \textbf{+25.8} & +9.2 \\
    Objective: LTR-2pass $\rightarrow$ CAP & M1/H16/O1/E8 & +0.2 & +24.0 & +5.0 & +8.8 & +9.5 \\
    Objective: LTR-1pass $\rightarrow$ CAP & M1/H16/O1/E16 & +1.0 & +20.4 & +13.8 & +23.4 & +14.7 \\
    $H:16\rightarrow8$ & CAP/M1/O1/E8 & -6.8 & +7.6 & +20.4 & +2.0 & +5.8 \\
    $O:1\rightarrow2$ & CAP/M1/H16/E8 & -10.4 & +6.6 & +5.6 & +6.2 & +2.0 \\
    $K:1\rightarrow0$ & CAP/M1/H16/O1/E8 & +0.2 & -12.8 & +9.8 & +3.6 & +0.2 \\
    $K:1\rightarrow8$ & CAP/M1/H16/O1/E8 & -2.2 & -10.4 & -4.0 & +3.8 & -3.2 \\
    $\lambda:.25\rightarrow.50$ & CAP/M1/H16/O1/E8 & -2.4 & -3.8 & +7.6 & +8.4 & +2.5 \\
    $M:1\rightarrow4$ & CAP/H16/O1/E8 & -1.8 & -2.0 & -3.8 & +9.8 & +0.6 \\
    $e:8\rightarrow16$ & CAP/M4/H16/O2 & -10.6 & -8.4 & +0.6 & -5.4 & -6.0 \\
    $e:8\rightarrow16$ & CAP/M1/H16/O1 & -3.6 & -6.6 & +7.8 & -0.6 & -0.8 \\
    $e:8\rightarrow16$ & LTR-1pass/M1/H16/O1 & -3.0 & -25.6 & +1.8 & +1.8 & -6.3 \\
    \midrule
    \multicolumn{7}{l}{\textit{Coupled configuration comparisons}} \\
    $H:8\rightarrow16,\ e:8\rightarrow16$ & CAP/M1/O1 & +3.2 & -14.2 & -12.6 & -2.6 & -6.5 \\
    $M:1\rightarrow4,\ H:8\rightarrow16,\ O:1\rightarrow2$ & CAP/E8 & +7.6 & -5.6 & -13.2 & +3.0 & -2.0 \\
    \bottomrule
  \end{tabular}
\end{table*}

\input{sections/appendix_a_table2_task_details}

\subsection{Targeted Single-Suite Diagnostics}
\label{app:targeted-libero-diagnostics}

The following experiments address narrow questions and therefore report only
the suite needed for the corresponding diagnosis.  This presentation keeps
their evidential scope explicit and separates them from the complete
four-suite matrix.

\begin{table}[!ht]
  \centering
  \caption{Execution-length ablation on LIBERO Long (\%).  Within each
  adjacent configuration block, training and model selection are fixed;
  only the number of open-loop executed actions $e$ changes.}
  \label{tab:long-execution-ablation}
  \tablebodyfont
  \setlength{\tabcolsep}{5pt}
  \begin{tabular}{llcc}
    \toprule
    Objective & Training configuration & $e$ & Long \\
    \midrule
    \method   & M1/H8/O1  & 4  & 30.4 \\
    \method   & M1/H8/O1  & 8  & 34.2 \\
    \midrule
    \method   & M1/H16/O1 & 8  & 32.2 \\
    \method   & M1/H16/O1 & 16 & 31.6 \\
    \midrule
    \method   & M4/H16/O2 & 4  & 35.2 \\
    \method   & M4/H16/O2 & 8  & 37.2 \\
    \method   & M4/H16/O2 & 16 & 31.8 \\
    \midrule
    LTR-1pass & M1/H16/O1 & 8  & 6.4 \\
    LTR-1pass & M1/H16/O1 & 16 & 8.2 \\
    \bottomrule
  \end{tabular}
\end{table}

For the \(H=8\) reference, executing the full predicted chunk gives the strongest
Long result among the tested values.  The two \(H=16\) policies show that the
preferred execution length remains task- and model-dependent: CAP is nearly
unchanged from \(e=8\) to \(e=16\) in the \(O=1\) family, while the
\(O=2\) policy degrades at \(e=16\).  These results motivate treating $e$ as a deployment
parameter tied to feedback frequency.

\paragraph{CAP gradient routing.}

\begin{table}[!ht]
  \centering
  \caption{CAP gradient-routing diagnostic on LIBERO Goal (\%).  Both rows
  use \(M=4,H=16,O=2,e=8\).  Decoder-only routing detaches VLM and resampler
  context for the CAP loss while preserving LTR gradients through all modules.}
  \label{tab:cap-gradient-routing}
  \tablebodyfont
  \begin{tabular}{lc}
    \toprule
    CAP gradient route & Goal \\
    \midrule
    VLM, resampler, and action decoder & 77.0 \\
    Action decoder only & 74.0 \\
    \bottomrule
  \end{tabular}
\end{table}

Blocking CAP gradients from the semantic context lowers Goal under the
matched \(M=4,H=16,O=2,e=8\) configuration.  A semantic-interference account
predicts the opposite direction of this drop; in this experiment, CAP benefits
from jointly shaping the shared context and action decoder.

\subsection{Suite-Resolved LIBERO-Plus Robustness}
\label{app:liberoplus-suite}

Tables~\ref{tab:liberoplus-suite}, \ref{tab:liberoplus-cap-h16-suite}, and
\ref{tab:liberoplus-ltr-suite} expand the aggregate robustness results in
Table~\ref{tab:liberoplus-main} by both suite and perturbation family.  Each
cell aggregates every instance assigned to its suite--factor block by the
released LIBERO-Plus task map.  Its Spatial, Object, Goal, and Long partitions
contain 2,402, 2,518, 2,591, and 2,519 instances, respectively, for 10,030 in
total.  We use this released partition throughout all controlled comparisons.

\begin{table*}[!t]
  \centering
  \caption{Suite-resolved zero-shot LIBERO-Plus success rate (\%) for the
  \(H=8,O=1,M=1\) \method policy.  ``Original'' is the corresponding
  standard-LIBERO
  score.  The All row averages Original over suites and aggregates each
  perturbation family over its official instances.}
  \label{tab:liberoplus-suite}
  \tablebodyfont
  \resizebox{\textwidth}{!}{%
  \begin{tabular}{lccccccccc}
    \toprule
    Suite & Original & Camera & Robot & Language & Light & Background & Noise & Layout & Plus mean \\
    \midrule
    Spatial & 88.6 & 25.5 & 50.6 & 84.6 & 92.8 & 78.3 & 32.8 & 76.9 & 61.9 \\
    Object  & 97.0 & 16.4 & 34.9 & 97.7 & 78.8 & 71.0 & 19.4 & 55.3 & 50.2 \\
    Goal    & 90.2 & 42.6 & 39.9 & 28.0 & 74.2 & 44.1 & 33.2 & 49.6 & 43.2 \\
    Long    & 34.2 &  8.8 & 13.5 & 19.1 &  6.6 & 12.5 &  3.8 & 13.1 & 10.9 \\
    \midrule
    All     & 77.5 & 23.3 & 34.3 & 56.2 & 63.9 & 50.0 & 21.2 & 50.6 & 41.3 \\
    \bottomrule
  \end{tabular}%
  }
\end{table*}

The suite matrix reveals strong task dependence within each perturbation
family.  Spatial and Object retain language-conditioned behavior well, while
Long degrades broadly under visual shifts.  Interpreting robustness therefore
requires both the perturbation factor and the task structure.  Failure-stage
annotations can further separate perception loss from long-horizon action
drift.

\begin{table*}[!t]
  \centering
  \caption{Suite-resolved zero-shot LIBERO-Plus success rate (\%) for the
  matched \(H=16,O=1,M=1\) \method policies deployed with $e=8$.  Every cell
  contains a released task-map suite--factor block.  The final row aggregates
  each factor over all four suites, and Plus mean aggregates the full
  10,030-instance perturbation partition.}
  \label{tab:liberoplus-cap-h16-suite}
  \tablebodyfont
  \resizebox{\textwidth}{!}{%
  \begin{tabular}{lccccccccc}
    \toprule
    Suite & Original & Camera & Robot & Language & Light & Background & Noise & Layout & Plus mean \\
    \midrule
    Spatial & 95.4 & 33.2 & 57.7 & 87.2 & 97.3 & 91.5 & 35.9 & 91.4 & 69.3 \\
    Object  & 89.4 & 23.2 & 26.9 & 93.2 & 73.4 & 70.6 & 27.0 & 56.1 & 50.1 \\
    Goal    & 69.8 & 42.2 & 27.6 & 36.1 & 52.3 & 29.2 & 41.4 & 33.9 & 37.1 \\
    Long    & 32.2 &  9.1 & 16.3 & 38.1 & 18.6 & 17.6 &  4.2 & 12.8 & 16.2 \\
    \midrule
    All     & 71.7 & 26.7 & 31.4 & 62.7 & 61.2 & 50.6 & 26.0 & 50.0 & 42.9 \\
    \bottomrule
  \end{tabular}%
  }
\end{table*}

The \(H=16\) matrix improves 20 of 28 suite--factor cells relative to
matched LTR-1pass, including every Language block, every Background block, and
all seven Long blocks.  Every factor also improves after four-suite
aggregation.  The result links CAP's in-distribution gains to robustness under
multiple observation and embodiment shifts.

\begin{table*}[!t]
  \centering
  \caption{Suite-resolved zero-shot LIBERO-Plus success rate (\%) for the
  \(H=16,O=1\) LTR-1pass policies deployed with $e=8$.  Every cell contains a
  released task-map suite--factor block.  The final row aggregates each factor
  over its four suite blocks, and Plus mean aggregates the full
  10,030-instance perturbation partition.}
  \label{tab:liberoplus-ltr-suite}
  \tablebodyfont
  \resizebox{\textwidth}{!}{%
  \begin{tabular}{lccccccccc}
    \toprule
    Suite & Original & Camera & Robot & Language & Light & Background & Noise & Layout & Plus mean \\
    \midrule
    Spatial & 93.8 & 33.8 & 49.4 & 75.6 & 89.0 & 89.9 & 37.9 & 81.8 & 63.9 \\
    Object  & 88.0 & 19.2 & 33.7 & 70.1 & 78.5 & 69.4 & 22.5 & 62.3 & 48.0 \\
    Goal    & 62.0 & 36.0 & 32.8 & 21.5 & 52.7 & 26.0 & 24.8 & 35.1 & 32.1 \\
    Long    &  6.4 &  1.2 &  4.1 &  4.4 &  3.3 &  4.5 &  0.7 &  3.8 &  3.0 \\
    \midrule
    All     & 62.6 & 22.2 & 29.5 & 42.2 & 56.8 & 45.5 & 20.3 & 47.7 & 36.4 \\
    \bottomrule
  \end{tabular}%
  }
\end{table*}

This LTR matrix supplies the suite--factor reference for the matched
\(H=16\) comparison.  It also shows that Spatial robustness can coexist with a
near-total Long collapse, which makes aggregate OOD scores insufficient for
diagnosing closed-loop policy quality.  The \(H=8\) CAP matrix provides broader
cross-horizon context, and the \(H=16\) CAP blocks provide the matched
comparison.

\begin{table*}[!t]
  \centering
  \caption{Suite-resolved zero-shot LIBERO-Plus success rate (\%) for the
  \(H=16,O=1\) LTR-2pass policies deployed with $e=8$.  Every entry is computed
  from a released task-map suite--factor block.  The All row pools the four
  suites within each perturbation family; its Plus mean aggregates the full
  10,030-instance perturbation partition.}
  \label{tab:liberoplus-ltr2-suite}
  \tablebodyfont
  \resizebox{\textwidth}{!}{%
  \begin{tabular}{lccccccccc}
    \toprule
    Suite & Original & Camera & Robot & Language & Light & Background & Noise & Layout & Plus mean \\
    \midrule
    Spatial & 95.2 & 29.3 & 51.4 & 83.6 & 92.1 & 85.3 & 34.2 & 83.1 & 64.3 \\
    Object  & 65.4 & 20.5 & 19.6 & 66.4 & 27.3 & 56.0 & 32.5 & 36.7 & 35.7 \\
    Goal    & 64.8 & 38.2 & 26.9 & 29.3 & 44.4 & 30.6 & 27.2 & 34.6 & 32.7 \\
    Long    & 23.4 &  5.5 &  8.7 & 23.0 &  5.5 & 13.8 &  3.6 &  8.0 &  9.6 \\
    \midrule
    All     & 62.2 & 23.1 & 25.9 & 50.0 & 42.8 & 45.1 & 23.5 & 42.0 & 35.2 \\
    \bottomrule
  \end{tabular}%
  }
\end{table*}

CAP improves 25 of the 28 suite--factor cells relative to this compute-matched
control, raises the Plus mean in every suite, and improves all seven pooled
perturbation families.  The Long advantage persists under distribution
shift, reinforcing the link between factorization diversity and extended
closed-loop robustness.

\subsection{Training Diagnostics}
\label{app:training-diagnostics}

Table~\ref{tab:training-diagnostics} records the final logging interval of the
\(H=16,O=2,M=4\) suite-specific experiments reported in
Table~\ref{tab:libero-completed-configs}.  These values are training-split
optimization diagnostics; closed-loop rollout success provides the
generalization metric and model assessment.  Negative NLL values are
valid densities under the learned continuous Gaussian, with suite-specific
scales that preclude classification-style cross-suite comparisons.

\begin{table*}[!t]
  \centering
  \caption{Final training-window diagnostics at step 30k for the
  \(H=16,O=2,M=4\) policies.  ``CAP NLL'' is the sampled
  anchored-permutation NLL.  L1, First L1, and End L1 are mean absolute
  action errors over all, first, and final chunk slots; Grip is gripper-state
  accuracy (\%).}
  \label{tab:training-diagnostics}
  \tablebodyfont
  \setlength{\tabcolsep}{5pt}
  \begin{tabular}{lcccccccc}
    \toprule
    Suite & Samples seen & LTR NLL & CAP NLL & L1 & First L1 & End L1 & Grip & Samples/s \\
    \midrule
    Spatial & 479,940 & -12.455 & -10.880 & 0.0451 & 0.1505 & 0.0421 & 98.28 & 6.33 \\
    Object  & 479,874 & -13.832 & -12.914 & 0.0386 & 0.1568 & 0.0333 & 98.09 & 6.47 \\
    Goal    & 479,900 & -14.759 & -13.837 & 0.0321 & 0.1272 & 0.0253 & 98.95 & 6.60 \\
    Long    & 479,960 & -14.462 & -13.635 & 0.0359 & 0.1581 & 0.0282 & 98.20 & 6.47 \\
    \bottomrule
  \end{tabular}
\end{table*}

\paragraph{Matched offline--closed-loop comparison.}
Table~\ref{tab:ltr-cap-long-diagnostics} compares the three matched
\(H=16,O=1,M=1\)
Long objectives after the same update and sample budget.  LTR-2pass preserves
the one-step training fit and improves closed-loop control; CAP improves
it further.  The ordering isolates a behavioral contribution from diverse
conditioning sets beyond repeated decoder computation.

\begin{table*}[!t]
  \centering
  \caption{Matched final training diagnostics and closed-loop Long success.
  All three objectives use \(H=16,O=1,M=1\), seed zero, 30k updates, and
  deterministic LTR deployment with $e=8$.  L1, First L1, and End L1 are mean
  absolute action errors over all, first, and final chunk slots.  ``Min-std''
  is the fraction of Gaussian log-scale outputs at the configured lower
  bound; Grip and Success are percentages.}
  \label{tab:ltr-cap-long-diagnostics}
  \tablebodyfont
  \setlength{\tabcolsep}{2.5pt}
  \begin{tabular}{lccccccccc}
    \toprule
    Training objective & LTR NLL & CAP NLL & L1 & First L1 & End L1 & Grip & Mean log-std & Min-std & Success \\
    \midrule
    LTR-1pass & -14.223 & n/a     & \textbf{0.0354} & \textbf{0.1607} & \textbf{0.0280} & 98.24 & -3.383 & 24.22 & 6.4 \\
    LTR-2pass & \textbf{-14.522} & n/a     & 0.0355 & 0.1617 & 0.0286 & 98.13 & -3.383 & 23.92 & 23.4 \\
    CAP       & -14.392 & -13.681 & 0.0361 & 0.1646 & 0.0282 & \textbf{98.32} & -3.379 & 23.87 & \textbf{32.2} \\
    \bottomrule
  \end{tabular}
\end{table*}

\subsection{Parameter Meanings and Sweep Semantics}
\label{app:parameter-semantics}

The symbols in the main ablation table affect different stages of training
and deployment.  Table~\ref{tab:parameter-semantics} fixes their meanings and
distinguishes the reference from the measured settings represented
in the study.

\begin{table*}[!t]
  \centering
  \caption{Meaning and experimental role of the principal PermVLA parameters.
  The final column lists settings represented by measured results.}
  \label{tab:parameter-semantics}
  \tablebodyfont
  \setlength{\tabcolsep}{5pt}
  \resizebox{\textwidth}{!}{%
  \begin{tabular}{clll}
    \toprule
    Symbol & Meaning & Reference & Values represented in the study \\
    \midrule
    $M$ & sampled CAP orders per context and update & 1 & 1, 4 \\
    $H$ & action steps predicted per policy query & 16 & 8, 16 \\
    $O$ & temporal observations supplied per camera view & 1 & 1, 2 \\
    $V$ & simultaneous RGB camera views per observation & 2 & 2 (primary + wrist; fixed) \\
    $e$ & leading predicted actions executed before replanning & 8 & 4, 8, 16 \\
    $K$ & fixed chronological prefix length in CAP orders & 1 & 0, 1, 8 (0 = unrestricted orders) \\
    $\lambda$ & weight of CAP in the mixed training objective & 0.25 & 0, 0.25, 0.50 (0 = LTR endpoint) \\
    CAP gradient & modules updated by the CAP loss & all modules & all modules; decoder only \\
    \bottomrule
  \end{tabular}
  }
\end{table*}

\subsection{Conditional Two-Stream Attention Implementation}
\label{app:two-stream}

We implement each reveal order with the conditional two-stream construction
introduced by XLNet \citep{yang2019xlnet}.  For physical slot \(i\), the
content and query streams start from
\begin{align}
  h_i^{(0)} &= E_a(a_{t+i-1}) + E_t(i), &
  g_i^{(0)} &= e_q + E_t(i),
  \label{eq:two-stream-inputs}
\end{align}
where \(E_a\) embeds an action, \(E_t\) identifies its physical time, and
\(e_q\) is a learned query token.  For sampled order \(z\), let \(r_z(i)\)
denote the reveal rank of slot \(i\).  Context tokens remain visible to every
slot, while action-to-action visibility is
\begin{align}
  M^{h}_{ij} &= \mathbb{I}[r_z(j) \leq r_z(i)], &
  M^{g}_{ij} &= \mathbb{I}[r_z(j) < r_z(i)].
  \label{eq:stream-masks}
\end{align}
The content stream can therefore read its own action and the previously
revealed actions; the query stream reads the strict revealed prefix.

At layer \(l\), both streams use the multimodal context and previous-layer
content states as their key-value source:
\begin{align}
  h_i^{(l)}
  &=
  \operatorname{Block}_h\!\left(
    h_i^{(l-1)},
    \context_t\cup
    \{h_j^{(l-1)}:r_z(j)\leq r_z(i)\}
  \right),\\
  g_i^{(l)}
  &=
  \operatorname{Block}_g\!\left(
    g_i^{(l-1)},
    \context_t\cup
    \{h_j^{(l-1)}:r_z(j)<r_z(i)\}
  \right).
  \label{eq:two-stream}
\end{align}
The two blocks share attention and feed-forward parameters.  Content states
provide the action key-value source, and the diagonal Gaussian head reads
\(g_i^{(L)}\).  Each prediction therefore conditions on exactly the strict
prefix defined by the sampled factorization.

\subsection{Experimental Parameters}
\label{app:complete-experimental-parameters}

Table~\ref{tab:configuration} consolidates the reference study's dataset,
architecture, optimization, evaluation, and hardware fields.  Diagnostic
variants are identified directly in the relevant rows.

\begin{table*}[!t]
  \centering
  \caption{Consolidated parameters for the controlled Gaussian
  PermVLA study.  Sweep rows in Table~\ref{tab:cap-controlled-sweep} replace
  one named value and keep every other entry fixed.}
  \label{tab:configuration}
  \tablebodyfont
  \setlength{\tabcolsep}{5pt}
  \renewcommand{\arraystretch}{0.96}
  \resizebox{\textwidth}{!}{%
  \begin{tabular}{lll}
    \toprule
    Category & Parameter & Value \\
    \midrule
    Data & Benchmark / suites & LIBERO Spatial, Object, Goal, Long; 10 tasks each \\
         & Demonstrations after no-op filtering & 432 / 454 / 428 / 379 \\
         & Input source & no-op-filtered Reinforcement Learning Datasets (RLDS) format; evaluation uses fixed benchmark episodes \\
         & Temporal stride / shuffle buffer & 1 / 10,000 \\
    \midrule
    Context & Backbone & Qwen3-VL-4B-Instruct \\
            & Adaptation & LoRA rank 32, alpha 16, dropout 0 \\
            & Images / views $V$ & primary + wrist RGB / 2, $256\times256$, train crop area 0.9 \\
            & Proprioception / history $O$ & 8D / 1 reference; $O=2$ diagnostic variant \\
            & Resampler & 32 tokens, width 768, 8 heads \\
    \midrule
    Action model & Decoder & conditional two-stream Transformer \\
                 & Width / layers & 768 / 8 \\
                 & Heads / dropout & 12 / 0.1 \\
                 & Action-side parameters & 67.0M \\
                 & Distribution & 7D diagonal Gaussian; log std in $[-4,1]$ \\
                 & Normalization & clipped 1\%/99\% training quantiles \\
    \midrule
    CAP & Reference $M,H,K,\lambda$ & 1 / 16 / 1 / 0.25 \\
        & Main objective & $0.75\mathcal L_{\mathrm{LTR}}+0.25\mathcal L_{\mathrm{CAP}}$ \\
        & LTR-1 / LTR-2 / CAP decoder passes & 1 / 2 / $1+M$ \\
        & Inference order & deterministic physical-time LTR \\
    \midrule
    PermDiffusion & Revealed-set construction & anchored order \(\pi\) with \(K=1\); cut \(m\sim\mathrm{Uniform}\{0,\ldots,H-1\}\) \\
        & Noise levels & revealed slots \(\tau=0\); targets share \(u\sim\mathrm{Uniform}\{0,\ldots,99\}\) \\
        & Objective & \(0.75\mathcal L_{\mathrm{DP}}+0.25\mathcal L_{\mathrm{PermDiff}}\) \\
        & Denoiser & full-self-attention Transformer; width 768, 8 layers, 12 heads \\
        & Per-slot time embedding & sinusoidal width 768 + MLP \(768\rightarrow3072\rightarrow768\) (SiLU) \\
        & Denoiser / LoRA learning rate & $10^{-4}$ / $10^{-5}$; 30k cosine schedule \\
        & Noise schedule & 100 corruption timesteps, cosine \(\bar\alpha\) \\
        & Inference & 10 synchronous deterministic DDIM steps (\(\eta=0\)), sample clamped to \([-1,1]\) \\
    \midrule
    Optimization & Optimizer / weight decay & AdamW / $10^{-4}$ \\
                 & Batch / accumulation / effective batch & 16 / 1 / 16 (reference); 8 / 2 / 16 (\(O=2\) protocol) \\
                 & Policy / LoRA learning rate & $10^{-4}$ / $10^{-5}$ \\
                 & Schedule & 30k cosine; 2,500 warmup; minimum $10^{-6}$ \\
                 & Gradient clipping & global norm 1.0 \\
                 & Precision / attention & bfloat16 / scaled dot-product attention (SDPA) \\
                 & Memory / compilation & gradient checkpointing disabled; compiled action policy \\
                 & Model snapshots / logging & every 10k / every 10 optimizer steps \\
    \midrule
    Deployment & Reference $O,H,e$ & 1 / 16 / 8 \\
               & LIBERO evaluation & 10 tasks $\times$ 50 episodes/suite with matched initial conditions \\
               & Success check & after every executed physical action \\
               & LIBERO-Plus & released task map; 10,030 frozen-policy instances; one rollout per instance \\
               & CALVIN & ABC$\rightarrow$D; \(H=8,O=1,e=8\); 100k updates; 1,000 five-instruction chains \\
    \midrule
    Hardware & Training topology & one single-GPU (NVIDIA A800 SMX4 80GB) process per suite-specific run \\
    \bottomrule
  \end{tabular}
  }
\end{table*}

\paragraph{CALVIN transfer protocol.}
The ABC$\rightarrow$D study trains all three objectives for 100k
updates on the same 17,870 ABC play-data episodes (1,071,743 frames) and seed
zero.  It uses Qwen3-VL-4B-Instruct with LoRA rank 64, primary and wrist RGB,
15D proprioception, relative 7D actions, \(H=8,O=1,e=8\), and deterministic LTR
deployment.  CAP uses $M=1$, $K=1$, and $\lambda=0.25$.  The effective batch
is 64 (batch 32 with two-step accumulation); policy and VLM learning rates are
$2\times10^{-4}$ and $2\times10^{-5}$ with 10k warmup steps and cosine decay.
Evaluation uses the official 1,000 fixed five-instruction chains in unseen
Scene D with at most 360 environment steps per instruction.

\paragraph{Result reporting.}
Every populated suite cell uses the full evaluation protocol and the same
initial-condition set described above.  External rows are transcribed from cited
sources for context, while matched CAP gains use the controlled study rows.

%% file: sections/appendix_a_table2_task_details.tex
\subsection{Task-Resolved Parameter Comparisons}
\label{app:table2-task-details}

Tables~\ref{tab:table2-task-spatial}--\ref{tab:table2-task-long} expand the
completed rows of Table~\ref{tab:cap-controlled-sweep}.  \emph{CAP ref} is the
absolute success rate of the \(M=1,H=16,O=1,e=8,K=1\) reference.  Every
remaining entry is the signed percentage-point change from that reference on
the same task.  LTR-1 and LTR-2 use \(e=8\); CAP/E16 and LTR/E16 denote the
two \(e=16\) deployment rows.  H8, O2, K0, and K8 change only the named parameter.
The \(\lambda=.50\) column changes only the CAP mixing weight.
Each task uses 50 episodes, and the suite-mean rows reproduce the corresponding
entries in Table~\ref{tab:cap-controlled-sweep}.

\begin{table*}[!ht]
  \centering
  \caption{Task-resolved LIBERO-Spatial comparison (\%).  Every task places
  the black bowl on the plate; the first column identifies its initial
  location.}
  \label{tab:table2-task-spatial}
  \tablebodyfont
  \setlength{\tabcolsep}{2.6pt}
  \renewcommand{\arraystretch}{0.96}
  \begin{tabular}{@{}p{0.25\textwidth}cccccccccc@{}}
    \toprule
    Task & CAP ref & \multicolumn{9}{c}{Change from CAP reference (percentage points)} \\
    \cmidrule(lr){3-11}
      & & LTR-1 & LTR-2 & H8 & O2 & CAP/E16 & LTR/E16 & K0 & K8 & \(\lambda=.50\) \\
    \midrule
    Between plate/ramekin & 100 & 0 & -2 & 0 & 0 & -2 & -2 & 0 & 0 & -2 \\
    Next to ramekin & 100 & -14 & -28 & -18 & -2 & -2 & -8 & -2 & -16 & -38 \\
    Table center & 100 & 0 & 0 & -2 & 0 & -4 & 0 & 0 & -2 & 0 \\
    On cookie box & 96 & +2 & +2 & -16 & +4 & +4 & -2 & 0 & +2 & -2 \\
    Top cabinet drawer & 80 & -2 & +10 & -4 & +2 & -18 & -16 & +8 & -16 & +10 \\
    On ramekin & 98 & -4 & +2 & 0 & 0 & -6 & -10 & -2 & -4 & 0 \\
    Next to cookie box & 94 & +4 & +4 & -18 & +4 & +4 & 0 & 0 & +4 & +4 \\
    On stove & 92 & 0 & +6 & 0 & -36 & -6 & +2 & +6 & +4 & +8 \\
    Next to plate & 98 & -6 & 0 & -2 & -4 & -6 & -10 & -6 & +2 & -2 \\
    On cabinet & 96 & +4 & +4 & -8 & -72 & 0 & 0 & -2 & +4 & -2 \\
    \midrule
    Suite mean & 95.4 & -1.6 & -0.2 & -6.8 & -10.4 & -3.6 & -4.6 & +0.2 & -2.2 & -2.4 \\
    \bottomrule
  \end{tabular}
\end{table*}

\begin{table*}[!ht]
  \centering
  \caption{Task-resolved LIBERO-Object comparison (\%).  Every task places
  the named object in the basket.}
  \label{tab:table2-task-object}
  \tablebodyfont
  \setlength{\tabcolsep}{2.6pt}
  \renewcommand{\arraystretch}{0.96}
  \begin{tabular}{@{}p{0.25\textwidth}cccccccccc@{}}
    \toprule
    Task & CAP ref & \multicolumn{9}{c}{Change from CAP reference (percentage points)} \\
    \cmidrule(lr){3-11}
      & & LTR-1 & LTR-2 & H8 & O2 & CAP/E16 & LTR/E16 & K0 & K8 & \(\lambda=.50\) \\
    \midrule
    Alphabet soup & 82 & +16 & -34 & +18 & +4 & -24 & -6 & 0 & +12 & -16 \\
    Cream cheese & 92 & 0 & -18 & +8 & +4 & +4 & +8 & -42 & -20 & +2 \\
    Salad dressing & 100 & -2 & -12 & 0 & 0 & 0 & -74 & -2 & -22 & 0 \\
    BBQ sauce & 52 & +14 & -26 & +44 & +34 & -10 & -42 & -34 & +18 & -20 \\
    Ketchup & 82 & 0 & -14 & +18 & +16 & -18 & -74 & -30 & -42 & +14 \\
    Tomato sauce & 96 & -12 & -68 & +4 & +2 & 0 & -30 & -10 & -24 & -4 \\
    Butter & 96 & +4 & -28 & -12 & +4 & +2 & -4 & -8 & -20 & +4 \\
    Milk & 96 & -30 & -16 & -6 & 0 & -16 & -42 & 0 & -4 & -20 \\
    Chocolate pudding & 98 & -2 & -20 & +2 & +2 & 0 & +2 & +2 & +2 & +2 \\
    Orange juice & 100 & -2 & -4 & 0 & 0 & -4 & -8 & -4 & -4 & 0 \\
    \midrule
    Suite mean & 89.4 & -1.4 & -24.0 & +7.6 & +6.6 & -6.6 & -27.0 & -12.8 & -10.4 & -3.8 \\
    \bottomrule
  \end{tabular}
\end{table*}

\begin{table*}[!ht]
  \centering
  \caption{Task-resolved LIBERO-Goal comparison (\%).  The first column gives
  a compact form of each language-conditioned goal.}
  \label{tab:table2-task-goal}
  \tablebodyfont
  \setlength{\tabcolsep}{2.6pt}
  \renewcommand{\arraystretch}{0.96}
  \begin{tabular}{@{}p{0.25\textwidth}cccccccccc@{}}
    \toprule
    Task & CAP ref & \multicolumn{9}{c}{Change from CAP reference (percentage points)} \\
    \cmidrule(lr){3-11}
      & & LTR-1 & LTR-2 & H8 & O2 & CAP/E16 & LTR/E16 & K0 & K8 & \(\lambda=.50\) \\
    \midrule
    Open middle drawer & 62 & -44 & -34 & +26 & -4 & +24 & -12 & -16 & -18 & +20 \\
    Bowl $\to$ stove & 98 & +2 & +2 & +2 & +2 & 0 & +2 & +2 & -6 & -4 \\
    Wine $\to$ cabinet & 98 & -8 & -40 & +2 & -10 & -4 & -16 & -8 & 0 & -8 \\
    Open top drawer + bowl & 2 & +72 & +62 & +70 & +46 & +36 & +50 & +50 & +30 & +44 \\
    Bowl $\to$ cabinet & 82 & +14 & +12 & +18 & +10 & +12 & +16 & +14 & 0 & +6 \\
    Push plate & 86 & -56 & +2 & +8 & -10 & -4 & -30 & +4 & +6 & -14 \\
    Cream cheese $\to$ bowl & 58 & -46 & -44 & +6 & +18 & +6 & -54 & +22 & -46 & +28 \\
    Turn on stove & 100 & -2 & -2 & 0 & -2 & 0 & -2 & 0 & 0 & 0 \\
    Bowl $\to$ plate & 96 & +4 & 0 & +4 & +4 & +2 & 0 & +4 & -4 & 0 \\
    Wine $\to$ rack & 16 & -14 & -8 & +68 & +2 & +6 & -14 & +26 & -2 & +4 \\
    \midrule
    Suite mean & 69.8 & -7.8 & -5.0 & +20.4 & +5.6 & +7.8 & -6.0 & +9.8 & -4.0 & +7.6 \\
    \bottomrule
  \end{tabular}
\end{table*}

\begin{table*}[!ht]
  \centering
  \caption{Task-resolved LIBERO-Long comparison (\%).  The first column
  summarizes each multi-stage instruction.}
  \label{tab:table2-task-long}
  \tablebodyfont
  \setlength{\tabcolsep}{2.6pt}
  \renewcommand{\arraystretch}{0.96}
  \begin{tabular}{@{}p{0.25\textwidth}cccccccccc@{}}
    \toprule
    Task & CAP ref & \multicolumn{9}{c}{Change from CAP reference (percentage points)} \\
    \cmidrule(lr){3-11}
      & & LTR-1 & LTR-2 & H8 & O2 & CAP/E16 & LTR/E16 & K0 & K8 & \(\lambda=.50\) \\
    \midrule
    Soup + sauce $\to$ basket & 32 & -32 & -16 & -30 & -30 & +6 & -30 & -12 & -14 & +16 \\
    Cream cheese + butter $\to$ basket & 30 & -30 & +2 & +18 & +6 & +2 & -24 & +56 & +8 & +38 \\
    Stove on + moka pot & 84 & -82 & -44 & -66 & -2 & -18 & -64 & -70 & +8 & -16 \\
    Bowl $\to$ drawer + close & 60 & -60 & -26 & +8 & -18 & -12 & -56 & -26 & +16 & -26 \\
    Two mugs $\to$ plates & 40 & -40 & -14 & +14 & -8 & -8 & -40 & +10 & 0 & +6 \\
    Book $\to$ caddy & 28 & +32 & -12 & -12 & +30 & -10 & +18 & +16 & +4 & +34 \\
    Mug + pudding placement & 4 & -4 & +16 & +14 & +48 & +34 & -4 & +36 & +16 & +16 \\
    Soup + cream cheese $\to$ basket & 34 & -34 & +2 & +40 & +16 & -20 & -30 & +12 & +2 & +10 \\
    Two moka pots $\to$ stove & 6 & -4 & -4 & +10 & +10 & +18 & -6 & +2 & 0 & +2 \\
    Mug $\to$ microwave + close & 4 & -4 & +8 & +24 & +10 & +2 & -4 & +12 & -2 & +4 \\
    \midrule
    Suite mean & 32.2 & -25.8 & -8.8 & +2.0 & +6.2 & -0.6 & -24.0 & +3.6 & +3.8 & +8.4 \\
    \bottomrule
  \end{tabular}
\end{table*}

The task-level changes show that aggregate effects are structured.  The O2
Spatial regression is concentrated in the stove and cabinet placements,
whereas its Object gain is broad.  Goal contains strong sign reversals across
instructions, so aggregate accuracy alone misses language- or
interaction-specific failures.  On Long, K8 improves several middle tasks
while weakening the first and final tasks; H8 and O2 obtain much of their gain
from tasks on which the reference remains weak.  The stronger CAP weight
improves six of ten Long tasks and several weak Goal tasks, while its Spatial
and Object losses are concentrated in a few placements and object identities.
K0 exposes a different trade-off: Object losses concentrate in cream cheese,
BBQ sauce, and ketchup; Goal gains concentrate in drawer insertion and two
placement tasks.  On Long, improved basket and placement tasks offset a sharp
loss on the stove-and-moka-pot task.  Similar suite means can therefore conceal
large task-level reversals.  These patterns motivate adapting prefix length
and regularization strength to the control problem.

%% file: sections/appendix_b_permdiffusion.tex
\section{Permutation-Regularized Diffusion Policies}
\label{app:permdiffusion}

This appendix develops a diffusion version of subset-conditioned
regularization
and reports a suite-level evaluation matrix together with a longer-horizon
reference experiment.  Diffusion action models provide a
natural representation of revealed subsets because every physical slot can
carry its own noise level.  We call this construction
\textsc{PermDiffusion}.

\subsection{Subset-Conditioned Diffusion Objective}
\label{app:permdiffusion-objective}

Let \(A=(a_1,\ldots,a_H)\) denote a physical-time action chunk and let
\(\context\) denote the multimodal context.  Standard diffusion policy
training samples a diffusion step \(u\), Gaussian noise
\(\epsilon\sim\mathcal N(0,I)\), and forms
\begin{equation}
  x_u = \sqrt{\bar\alpha_u}A
      + \sqrt{1-\bar\alpha_u}\epsilon .
\end{equation}
An epsilon-prediction denoiser maps a hybrid chunk \(\widetilde A\), a
per-slot diffusion-step vector \(\tau\), a binary reveal mask
\(\mathbf 1_S\), and the context to per-slot noise predictions
\(\epsilon_{\theta,i}(\widetilde A,\tau,\mathbf 1_S,\context)\).  Standard
diffusion policy training is the special case with no revealed slots and a
uniform diffusion step, minimizing
\begin{equation}
  \mathcal L_{\mathrm{DP}}
  = \mathbb E_{A,u,\epsilon}
    \left[\left\|\epsilon-\epsilon_\theta(x_u,u\mathbf 1,\mathbf
    0,\context)\right\|_2^2\right].
  \label{eq:appendix-dp-loss}
\end{equation}

For the permutation term, we sample \(\pi\sim\Pi_K\), using \(K=1\) in the
reported experiments, and a cut position \(m\) uniformly from
\(\{0,\ldots,H-1\}\), so the target set is never empty.  The revealed set is
\(S=\{\pi_1,\ldots,\pi_m\}\), and \(U=\{1,\ldots,H\}\setminus S\) contains
the prediction targets; \(m=0\) recovers fully noisy unconditional denoising.
Slots in \(S\) receive clean expert actions and a
zero diffusion step.  Slots in \(U\) receive their noisy actions and the
sampled diffusion step \(u\).  With \(\widetilde A\) denoting this hybrid
chunk, the auxiliary loss is
\begin{equation}
  \mathcal L_{\mathrm{PermDiff}}
  = \mathbb E
    \left[
      \frac{1}{|U|}\sum_{i\in U}
      \left\|\epsilon_i-
      \epsilon_{\theta,i}(\widetilde A,\tau,\mathbf 1_S,\context)\right\|_2^2
    \right],
  \label{eq:permdiff-loss}
\end{equation}
where \(\tau_i=0\) for \(i\in S\), \(\tau_i=u\) for \(i\in U\), and
\(\mathbf 1_S\) identifies revealed slots.  The physical positions of the
actions remain fixed; the permutation determines the revealed subset.

The prototype jointly optimizes
\begin{equation}
  \mathcal L_{\mathrm{PermDiffusion}}
  = 0.75\mathcal L_{\mathrm{DP}}
  + 0.25\mathcal L_{\mathrm{PermDiff}} .
  \label{eq:permdiff-total}
\end{equation}
This construction trains one denoiser to solve unconditional chunk denoising
and clean-subset-conditioned completion.

\subsection{Architecture, Training, and Sampling}
\label{app:permdiffusion-implementation}

\begin{figure*}[!t]
  \centering
  \input{figures/permdiffusion_architecture}
  \caption{\textsc{PermDiffusion} architecture.  During training, a sampled
  anchored order and cut divide the expert chunk into clean revealed slots
  \(S\) and noisy prediction slots \(U\).  Physical positions, per-slot noise
  levels, and revealed-mask embeddings identify every input to the shared
  denoiser.  Standard diffusion is the special case in which all slots are
  noisy.  Deployment begins from Gaussian noise and applies ordinary
  synchronous DDIM updates to the full physical-time chunk; the sampled
  subset construction is used only for training.}
  \label{fig:permdiffusion-architecture}
\end{figure*}
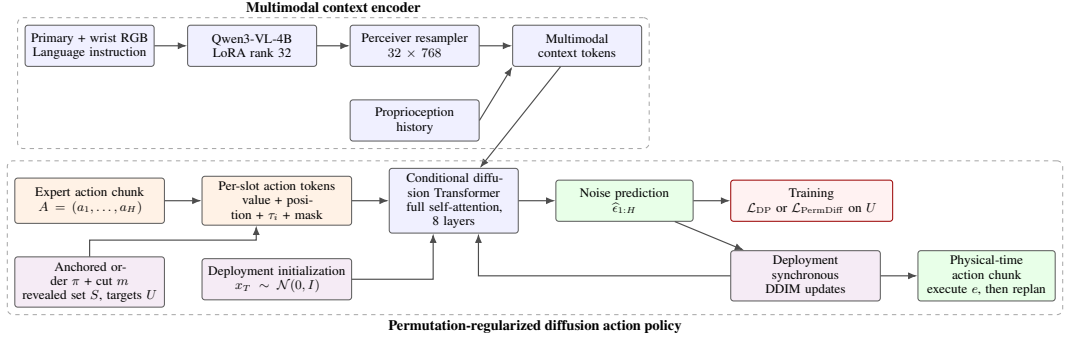

Figure~\ref{fig:permdiffusion-architecture} summarizes the shared context
encoder, subset-conditioned training path, and standard deployment path.
\textsc{PermDiffusion} reuses the Qwen3-VL context encoder, multimodal
resampler, action normalization, training data pipeline, and benchmark
evaluation interface from \method.  Its action module is a
full-self-attention Transformer denoiser.
Each action slot combines a projected noisy action, a physical-position
embedding, a sinusoidal per-slot diffusion-time embedding processed by a
multilayer perceptron (MLP), and a learned revealed-mask embedding.  The
action stream cross-attends
to the shared visual, language, and proprioceptive context.

The suite-level configuration uses the same Qwen3-VL encoder and multimodal
resampler, an \(H=8,O=1,e=8\) temporal configuration, a 768-wide
eight-layer denoiser, and 30k optimizer updates.  A separate Goal reference
uses \(H=16,O=2,e=8\) with
the same model width, objective, optimizer budget, and data protocol.  Both
configurations use 100 diffusion corruption timesteps with a cosine noise
schedule and ten synchronous deterministic
denoising diffusion implicit model (DDIM) inference steps.  Deployment starts
from a Gaussian action chunk,
denoises all slots at every update, and executes the first eight physical
actions before replanning.  The permutation objective therefore changes
training constraints while preserving the standard diffusion sampling path.

\subsection{LIBERO Suite Results}
\label{app:permdiffusion-results}

Table~\ref{tab:permdiffusion-results} reports the \(H=8,O=1,e=8\)
suite matrix and a standalone \(H=16,O=2,e=8\) Goal reference.  Each
populated entry uses 50 episodes for each of ten tasks.  The Goal reference
shows that the construction also trains successfully at a longer horizon;
the suite-level comparison uses the temporally matched \(H=8,O=1,e=8\)
configuration.

\begin{table*}[!t]
  \centering
  \caption{\textsc{PermDiffusion} results on LIBERO.  The \(H=8,O=1,e=8\) rows form
  the suite matrix, while the \(H=16,O=2,e=8\) Goal row is a standalone
  reference.  Each entry is a 500-episode success rate (\%).}
  \label{tab:permdiffusion-results}
  \tablebodyfont
  \setlength{\tabcolsep}{8pt}
  \begin{tabular}{lccc}
    \toprule
    Treatment & Temporal config. & Suite & Success \\
    \midrule
    \textsc{PermDiffusion} & H8/O1/E8 & Spatial & 87.6 \\
    \textsc{PermDiffusion} & H8/O1/E8 & Object & 89.8 \\
    \textsc{PermDiffusion} & H8/O1/E8 & Goal & 84.2 \\
    \textsc{PermDiffusion} & H8/O1/E8 & Long & 36.8 \\
    \midrule
    \textsc{PermDiffusion} & H16/O2/E8 & Goal & 82.2 \\
    \bottomrule
  \end{tabular}
\end{table*}

The complete \(H=8\) matrix averages 74.6\% and shows that the construction
supports spatial, object, goal-conditioned, and multi-stage control while
retaining standard DDIM sampling.  Relative to the temporally matched CAP
reference, the diffusion prototype is close on Spatial, lower on Object and
Goal, and higher on Long.  These results establish the feasibility of
subset-conditioned regularization beyond autoregressive likelihoods; isolating
the benefit of the auxiliary objective requires a matched \(\lambda=0\)
diffusion control.  The different
suite profile also suggests that useful subset structures depend on the action
generator.  Matched one-pass and compute-matched diffusion controls can next
isolate the contribution of subset-conditioned denoising and reveal how
the revealed-subset structure interacts with the corruption schedule.

%% file: figures/permdiffusion_architecture.tex
\resizebox{\textwidth}{!}{%
\begin{tikzpicture}[
  font=\small,
  >=Latex,
  line/.style={-Latex, thick, draw=black!70},
  contextblock/.style={
    draw=black!65,
    rounded corners=2pt,
    fill=blue!6,
    align=center,
    minimum height=10mm,
    text width=29mm
  },
  trainblock/.style={
    draw=black!65,
    rounded corners=2pt,
    fill=orange!10,
    align=center,
    minimum height=11mm,
    text width=34mm
  },
  control/.style={
    draw=black!65,
    rounded corners=2pt,
    fill=violet!8,
    align=center,
    minimum height=10mm,
    text width=34mm
  },
  output/.style={
    draw=black!65,
    rounded corners=2pt,
    fill=green!9,
    align=center,
    minimum height=10mm,
    text width=31mm
  },
  loss/.style={
    draw=red!60!black,
    rounded corners=2pt,
    fill=red!5,
    align=center,
    minimum height=10mm,
    text width=37mm
  },
  group/.style={
    draw=black!35,
    dashed,
    rounded corners=4pt,
    inner sep=5pt
  }
]

\node[contextblock] (obs) {Primary + wrist RGB\\Language instruction};
\node[contextblock, right=8mm of obs] (qwen) {Qwen3-VL-4B\\LoRA rank 32};
\node[contextblock, right=8mm of qwen] (resampler)
  {Perceiver resampler\\\(32\times768\)};
\node[contextblock, right=8mm of resampler] (context)
  {Multimodal\\context tokens};
\node[contextblock, below=8mm of resampler] (proprio)
  {Proprioception\\history};

\draw[line] (obs) -- (qwen);
\draw[line] (qwen) -- (resampler);
\draw[line] (resampler) -- (context);
\draw[line] (proprio.east) -- (context.south west);

\node[trainblock, below=27mm of obs] (expert)
  {Expert action chunk\\\(A=(a_1,\ldots,a_H)\)};
\node[control, below=8mm of expert] (subset)
  {Anchored order \(\pi\) + cut \(m\)\\revealed set \(S\), targets \(U\)};
\node[trainblock, right=9mm of expert] (tokens)
  {Per-slot action tokens\\value + position + \(\tau_i\) + mask};
\node[contextblock, right=9mm of tokens] (denoiser)
  {Conditional diffusion Transformer\\full self-attention, 8 layers};
\node[output, right=9mm of denoiser] (prediction)
  {Noise prediction\\\(\widehat\epsilon_{1:H}\)};

\node[loss, right=9mm of prediction] (objective)
  {Training\\\(\mathcal L_{\mathrm{DP}}\) or
   \(\mathcal L_{\mathrm{PermDiff}}\) on \(U\)};
\node[control, below right=7mm and 9mm of prediction] (ddim)
  {Deployment\\synchronous DDIM updates};
\node[output, right=9mm of ddim] (chunk)
  {Physical-time action chunk\\execute \(e\), then replan};
\node[control, below=8mm of tokens] (initial)
  {Deployment initialization\\\(x_T\sim\mathcal N(0,I)\)};

\draw[line] (expert) -- (tokens);
\draw[line] (subset.north) -- ++(0,4mm)
  -| ($(tokens.south)+(-5mm,0)$);
\draw[line] (tokens) -- (denoiser);
\draw[line] (context) -- (denoiser);
\draw[line] (denoiser) -- (prediction);
\draw[line] (prediction) -- (objective);
\draw[line] (prediction) -- (ddim);
\draw[line] (initial.east) -| ($(denoiser.south)+(-5mm,0)$);
\draw[line] (ddim.west) -| ($(denoiser.south)+(6mm,0)$);
\draw[line] (ddim) -- (chunk);

\node[group, fit=(obs)(qwen)(resampler)(context)(proprio),
      label={[font=\bfseries]above:Multimodal context encoder}] {};
\node[group, fit=(expert)(subset)(tokens)(initial)(denoiser)(prediction)
                       (objective)(ddim)(chunk),
      label={[font=\bfseries]below:Permutation-regularized diffusion action policy}] {};

\end{tikzpicture}%
}

%% file: sections/appendix_c_theory.tex
\section{Theoretical Interpretation and Testable Hypotheses}
\label{app:cap-theory}

\begin{figure*}[t]
  \centering
  \input{figures/loss_overview}
  \caption{Loss construction in \method.  Standard LTR training follows one
  physical-time factorization and sums teacher-forced per-slot likelihoods
  against the expert chunk.  CAP samples factorization orders over the same
  physical action variables, illustrated here with \(K=1\); each order defines
  revealed subsets and a corresponding likelihood.  Their Monte Carlo average
  is a training-only regularizer of the shared policy.  The drawn losses are
  unnormalized within-chunk sums; Equations~\ref{eq:ltr-loss}--\ref{eq:cap-loss}
  define the normalized objectives.  Action indices denote
  fixed physical slots in both panels, and deployment retains deterministic
  physical-time LTR control.}
  \label{fig:loss-overview}
\end{figure*}
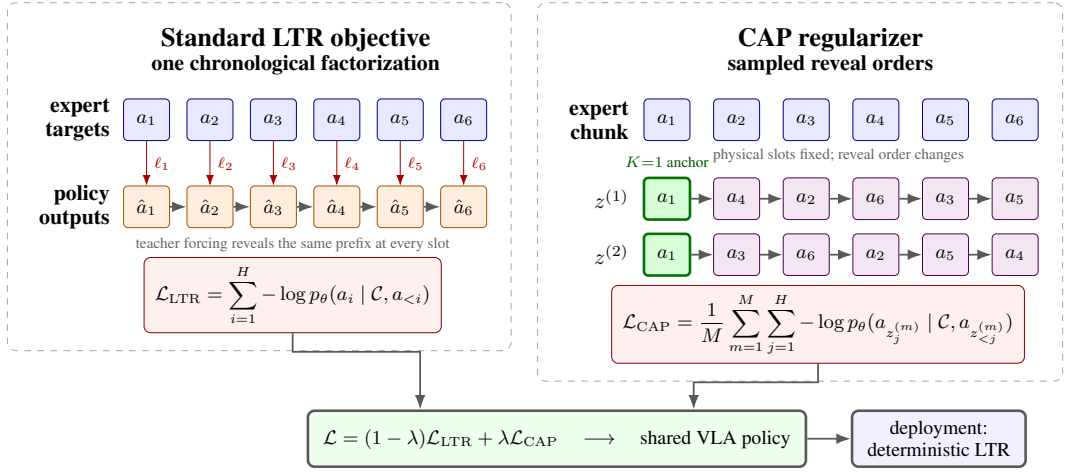

This appendix separates an exact chain-rule result from hypotheses about
finite data, shared parameters, optimization, and closed-loop control.  The
identity establishes that valid factorization orders describe the same expert
joint distribution under realizability.  The hypotheses explain why training
one finite policy across those orders can change its representation and
generalization.  Figure~\ref{fig:loss-overview} contrasts the two losses
analyzed below: the single teacher-forced path of standard LTR training and
the order-averaged CAP objective over the same physical action variables.

\begin{figure*}[h]
  \centering
  \includegraphics[width=0.8\linewidth]{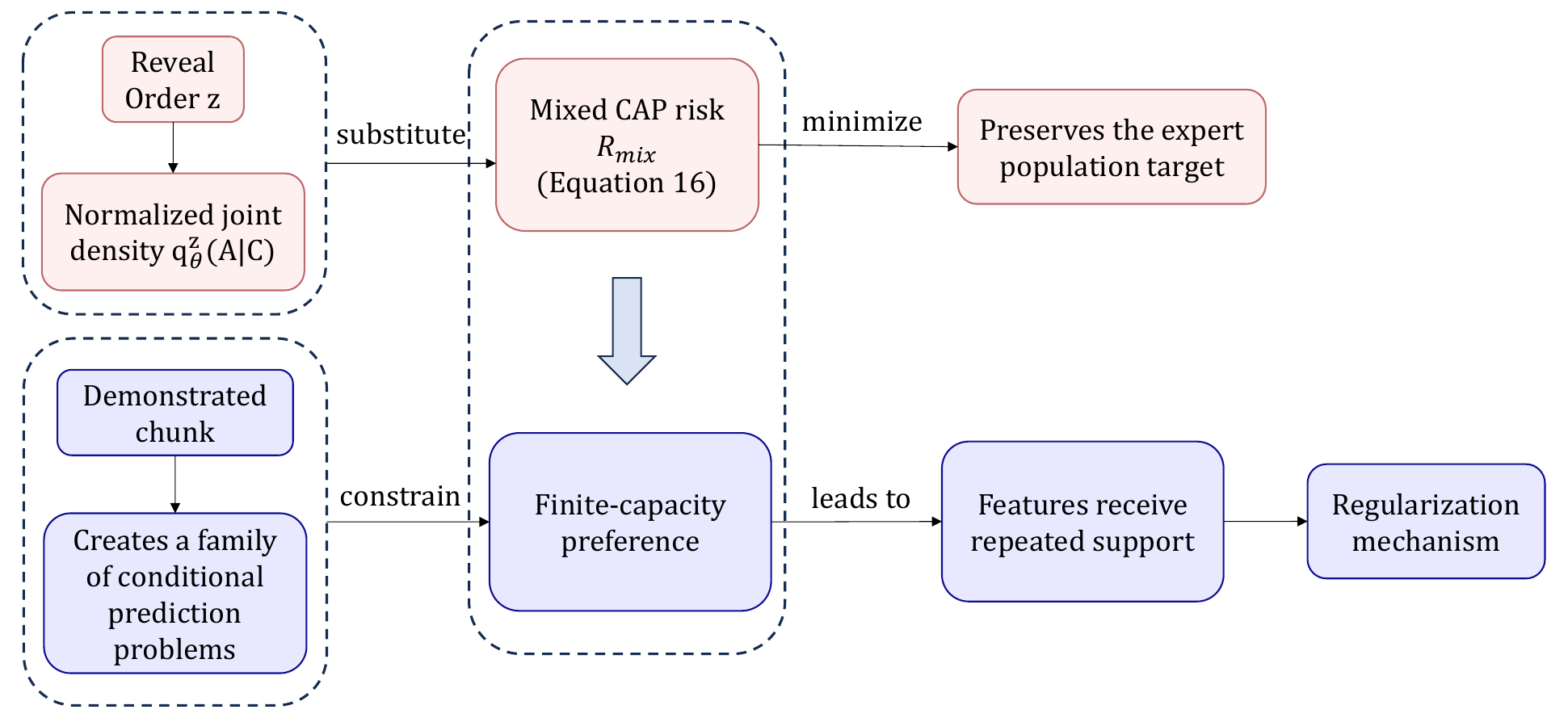}
  \caption{\textbf{Illustration of CAP's theoretical interpretation.}
In the upper path, a reveal order \(z\) defines a normalized joint density
\(q_\theta^z(\actions\mid\context)\); substituting this density into the mixed
CAP risk \(\mathcal R_{\mathrm{mix}}\) (Equation~\ref{eq:cap-mixed-population})
and minimizing it preserves the expert population target. In the lower path, a demonstrated action chunk creates a family of conditional prediction
problems; these problems jointly constrain a finite-capacity model, so features
that remain predictive across conditional sets receive repeated support,
thereby forming CAP's regularization mechanism.}
  \label{fig:C1_C2_explanation}
\end{figure*}

\subsection{Exact Chain-Rule Result Across Reveal Orders}

Let \(p^*(\actions\mid\context)\) denote the expert conditional distribution
and let \(q_\theta^z(\actions\mid\context)\) be the normalized joint density
induced by the shared decoder under reveal order \(z\).  The population
negative log likelihood for every valid order satisfies
\begin{equation}
  \begin{aligned}
  \mathcal R_z(\theta)
  &=
  \mathbb E_{p^*}
  \left[-\log q_\theta^z(\actions\mid\context)\right] \\
  &=
  H_{p^*}(\actions\mid\context)
  +
  \mathbb E_{\context}
  \mathrm{KL}\!\left(
    p^*(\actions\mid\context)
    \,\|\,
    q_\theta^z(\actions\mid\context)
  \right).
  \end{aligned}
  \label{eq:cap-kl}
\end{equation}
Here \(H_{p^*}(\actions\mid\context)\) denotes the conditional differential
entropy of the continuous action chunk; the minimizer is independent of its
value.  The mixed CAP population risk is therefore
\begin{equation}
  \begin{aligned}
  \mathcal R_{\mathrm{mix}}(\theta)
  ={}& H_{p^*}(\actions\mid\context) \\
  &+(1-\lambda)
  \mathbb E_{\context}\mathrm{KL}
  \!\left(p^*\,\|\,q_\theta^{\mathrm{LTR}}\right)
  +\lambda\,
  \mathbb E_{z\sim\Pi_K}\mathbb E_{\context}
  \mathrm{KL}\!\left(p^*\,\|\,q_\theta^z\right).
  \end{aligned}
  \label{eq:cap-mixed-population}
\end{equation}
Under realizability, one shared parameter setting represents \(p^*\) under
every order and minimizes every term.  CAP consequently preserves the expert
population target while changing the finite-model optimization geometry.
These identities hold for every \(K\in\{0,\ldots,H\}\), including the
unrestricted family \(\Pi_0\).  Prefix length selects the auxiliary
factorizations and their training bias; the retained LTR objective and
chronological deployment specify the control path.

\subsection{Finite-Capacity Regularization Through Conditional Sets}

Practical VLA training uses finite demonstrations, restricted model capacity,
shared features, and stochastic optimization.  These conditions couple the
factorization tasks.  LTR constrains action \(a_i\) under one prefix family
\(a_{<i}\).  CAP also constrains the same physical action under sampled
subsets selected by the reveal order.  Each demonstrated chunk therefore
creates a family of conditional prediction problems while preserving its
physical target trajectory.

We interpret this construction as \emph{conditional-set augmentation}.  A
feature that predicts an action only through one teacher-forced chain receives
limited support across the sampled tasks.  Features encoding relations that
remain predictive under several valid revealed subsets receive repeated
support through the shared decoder and context representation.  This creates
a finite-capacity preference among models that can fit the same LTR training
examples.  The preference is the proposed regularization mechanism.

For independent samples \(z_1,\ldots,z_M\sim\Pi_K\), the per-example CAP
gradient estimator is
\begin{equation}
  \widehat g_M
  = \frac{1}{M}\sum_{m=1}^{M}
  \nabla_\theta \mathcal L_{z_m},
  \qquad
  \mathbb E[\widehat g_M]=
  \mathbb E_{z\sim\Pi_K}[\nabla_\theta\mathcal L_z].
  \label{eq:cap-gradient-estimator}
\end{equation}
Conditioned on one expert chunk, independent order sampling gives
\begin{equation}
  \mathrm{Var}(\widehat g_M\mid\context,\actions)
  = \frac{1}{M}\mathrm{Var}_{z\sim\Pi_K}
  \!\left(\nabla_\theta\mathcal L_z\mid\context,\actions\right),
  \label{eq:cap-gradient-variance}
\end{equation}
where \(\mathrm{Var}\) denotes per-coordinate variance, so the scalar
second-moment reduction is \(1/M\) for the trace of the gradient covariance.
Larger \(M\) thus improves estimation of the order-averaged gradient.  It also
changes decoder computation and the relative frequency with which one chunk
constrains the shared parameters.  These two effects motivate the
compute-matched LTR-2pass control and the targeted \(M\) diagnostic.

\subsection{Factorization Sensitivity as a Diagnostic}

The factorization-regularization hypothesis can be tested by comparing the joint
densities induced by different reveal orders.  Equation
\ref{eq:factorization-sensitivity} defines factorization sensitivity as the
squared discrepancy between the joint log likelihoods induced by two reveal
orders on the same expert chunk.
CAP optimizes average likelihood over orders; \(\mathcal S\) is an external
diagnostic of agreement between reveal orders.  Shared parameters and
common expert targets suggest that successful CAP training may reduce this
sensitivity across expert chunks.  We test this prediction using the matched
\(M=1,H=16,O=1\) models.  For scale-independent reporting, we define
\(\mathcal S_H=\mathcal S/H^2\), which compares per-step sequence log
likelihoods.  Each suite uses a deterministic, task-balanced reservoir of 64
training chunks per task (640 total) and the same 16 independently sampled
orders from \(\Pi_1\) for LTR-1pass, compute-matched LTR-2pass, and CAP.  We
also report the per-step NLL increase under other reveal orders,
\(\widetilde{\Delta}_{\mathrm{arb}}=\operatorname{median}_{(\context,\actions)}
[\mathbb E_z\mathcal L_z/H-\mathcal L_{\mathrm{LTR}}/H]\).  The training
protocol uses all demonstrations, so this experiment diagnoses the learned
factorization behavior on a fixed training subset; LIBERO-Plus supplies the
separate OOD test.

\begin{table*}[!t]
  \centering
  \caption{Factorization sensitivity of matched \(M=1,H=16,O=1\) models.
  Tildes denote medians over 640 expert chunks per suite; brackets give the
  interquartile range of \(\mathcal S_H\).  Each chunk is scored under the
  same 16 uniformly sampled, causally anchored orders with \(K=1\).
  \(\widetilde{\Delta}_{\mathrm{arb}}\) is arbitrary-order minus LTR NLL per
  action slot.
  Lower sensitivity and gap indicate greater agreement across factorization orders.}
  \label{tab:factorization-sensitivity}
  \tablebodyfont
  \setlength{\tabcolsep}{5pt}
  \renewcommand{\arraystretch}{0.98}
  \begin{tabular}{llccc}
    \toprule
    Suite & Training objective &
    \(\widetilde{\mathcal S}_H\) [IQR] &
    \(\widetilde{\Delta}_{\mathrm{arb}}\) &
    LTR NLL / slot \\
    \midrule
    Long
      & LTR-1pass & 384.85 [114.00, 1{,}473.49] & 59.11 & -15.16 \\
      & LTR-2pass & 407.21 [110.73, 1{,}570.34] & 59.98 & -15.18 \\
      & CAP & \textbf{0.108 [0.055, 0.215]} & \textbf{0.97} & -15.16 \\
    \midrule
    Goal
      & LTR-1pass & 787.74 [177.30, 7{,}161.38] & 80.36 & -15.34 \\
      & LTR-2pass & 1{,}049.71 [216.04, 9{,}728.23] & 84.91 & -15.42 \\
      & CAP & \textbf{0.113 [0.062, 0.223]} & \textbf{1.09} & -15.28 \\
    \midrule
    Spatial
      & LTR-1pass & 1{,}214.51 [282.81, 16{,}174.20] & 89.83 & -13.06 \\
      & LTR-2pass & 1{,}160.07 [286.16, 13{,}469.60] & 91.96 & -13.18 \\
      & CAP & \textbf{0.124 [0.071, 0.219]} & \textbf{0.94} & -13.07 \\
    \midrule
    Object
      & LTR-1pass & 1{,}436.65 [226.10, 11{,}003.02] & 98.36 & -14.38 \\
      & LTR-2pass & 1{,}015.26 [185.05, 8{,}609.91] & 87.84 & -14.47 \\
      & CAP & \textbf{0.128 [0.059, 0.240]} & \textbf{0.86} & -14.45 \\
    \bottomrule
  \end{tabular}
\end{table*}

Table~\ref{tab:factorization-sensitivity} shows the same diagnostic pattern
in every suite.  CAP reduces median sensitivity between reveal orders by several
orders of magnitude and reduces the corresponding NLL increase from tens of nats
per slot to approximately one.  Its standard LTR NLL remains closely matched
to both controls.  A second LTR pass leaves agreement between reveal orders and
robustness to unseen factorization orders unchanged.  CAP changes the learned
factorization behavior through order-diverse supervision while retaining the
fit under the LTR factorization.  Together with the closed-loop and
LIBERO-Plus results,
this observation connects the intended training pressure to the learned
behavior across reveal orders; causal mediation between agreement and OOD success is a
promising direction for intervention-based evaluation.

We next intervene on the CAP weight, the number of sampled orders, and the
anchor.  Every row in Table~\ref{tab:factorization-interventions} uses the
same task-balanced chunks and the same \(\Pi_1\) evaluation orders.  The
\(\lambda\) rows form a direct strength sweep, while \(M\) changes Monte Carlo
coverage and \(K\) changes the support of the training order family.

\begin{table*}[!t]
  \centering
  \caption{Factorization-sensitivity interventions on Goal and Long.
  \(\widetilde{\mathcal S}_H\) and \(\widetilde{\Delta}_{\mathrm{arb}}\)
  are medians under the same 16 \(\Pi_1\) evaluation orders; success is the
  corresponding 500-episode LIBERO result (\%).  LTR is the
  \(\lambda=0\) endpoint.}
  \label{tab:factorization-interventions}
  \tablebodyfont
  \setlength{\tabcolsep}{5pt}
  \begin{tabular}{llccccc}
    \toprule
    Suite & Training setting & \(\lambda\) & \(M\) & \(K\) &
    \(\widetilde{\mathcal S}_H\) &
    \(\widetilde{\Delta}_{\mathrm{arb}}\) / Success \\
    \midrule
    Long & LTR-1pass       & 0    & -- & -- & 384.850 & 59.11 / 6.4 \\
         & CAP reference  & 0.25 & 1  & 1  & 0.108   & 0.97 / 32.2 \\
         & CAP weight     & 0.50 & 1  & 1  & 0.079   & 0.66 / 40.6 \\
         & CAP orders     & 0.25 & 4  & 1  & 0.103   & 0.93 / 42.0 \\
         & Anchor         & 0.25 & 1  & 8  & 182.307 & 29.09 / 36.0 \\
    \midrule
    Goal & LTR-1pass      & 0    & -- & -- & 787.740 & 80.36 / 62.0 \\
         & CAP reference & 0.25 & 1  & 1  & 0.113   & 1.09 / 69.8 \\
         & CAP weight    & 0.50 & 1  & 1  & 0.083   & 0.75 / 77.4 \\
         & CAP orders    & 0.25 & 4  & 1  & 0.111   & 1.09 / 66.0 \\
         & Anchor        & 0.25 & 1  & 8  & 236.312 & 32.66 / 65.8 \\
    \bottomrule
  \end{tabular}
\end{table*}

The weight sweep provides a dose-response signature: stronger CAP pressure
reduces reveal-order disagreement and raises success on both suites.
Increasing \(M\) instead leaves \(\widetilde{\mathcal S}_H\) close to the
reference while improving Long, so its effect extends beyond this
single global statistic.  The \(K=8\) row is evaluated outside its narrower
training family; its three-orders-of-magnitude sensitivity increase coincides
with \emph{higher} Long success, and conditional independence among the action
slots would minimize \(\widetilde{\mathcal S}_H\) trivially without implying
coordinated execution.  The diagnostic therefore verifies agreement over the
trained permutation family, with control capability assessed separately by
closed-loop success.
Table~\ref{tab:anchor-family-cross-eval} separates training
coverage from evaluation order.

\begin{table}[!t]
  \centering
  \caption{Anchor-family cross-evaluation.  Entries are medians
  \(\widetilde{\mathcal S}_H\) over the same chunks.  Training with \(K=1\)
  covers the broad
  \(\Pi_1\) family; \(K=8\) covers the narrower \(\Pi_8\) family.}
  \label{tab:anchor-family-cross-eval}
  \tablebodyfont
  \setlength{\tabcolsep}{7pt}
  \begin{tabular}{lccc}
    \toprule
    Suite & Training family & Evaluate \(\Pi_1\) & Evaluate \(\Pi_8\) \\
    \midrule
    Long & \(\Pi_1\) & 0.108 & 0.055 \\
         & \(\Pi_8\) & 182.307 & 0.038 \\
    \midrule
    Goal & \(\Pi_1\) & 0.113 & 0.061 \\
         & \(\Pi_8\) & 236.312 & 0.051 \\
    \bottomrule
  \end{tabular}
\end{table}

Both policies agree across orders drawn from their respective training
families.  The broadly trained \(\Pi_1\) policy also remains consistent on
the nested \(\Pi_8\) family, while the \(\Pi_8\) policy becomes sensitive when
evaluated on the unseen order support in \(\Pi_1\).  The learned invariance
therefore follows the conditioning subsets supplied by CAP training.  This
support-dependent signature is direct evidence for conditional-set
augmentation and defines the scope of the mechanism.

The Long reversal between pointwise action error and closed-loop success is
consistent with improved coherent continuation, calibration, or recovery
under states induced by the policy.  Action error without teacher forcing,
perturb-and-recover rollouts, trajectory-level likelihood variance, and
success-conditioned uncertainty can distinguish these explanations.
Layer-wise probes and interventions on revealed actions can then identify the
conditional dependencies across time that drive successful continuation.

\subsection{Connection to OOD Generalization}

Consider an environment shift that changes visual appearance, language form,
or embodiment while preserving part of the conditional structure among
future expert actions.  A policy specialized to one teacher-forced path can
couple its predictions to context features that are reliable on the training
distribution and fragile after the shift.  Conditional-set augmentation asks
the shared representation to support several valid action-completion views.
We hypothesize that this pressure favors conditional dependencies that remain useful
when some contextual correlations change.

This interpretation predicts structured OOD behavior.  CAP should help most
when the shift preserves conditional dependencies within an action chunk and disrupts
context correlations tied to the LTR factorization.  Its advantage can shrink
when the shift
changes the action semantics themselves, removes information required by all
conditional views, or makes generated action prefixes unreliable.  The broad
LIBERO-Plus gains support the first case; the suite-resolved variation
provides targets for perception, language, and embodiment probes.

This view also motivates a stricter evaluation protocol for VLA
regularization.  An in-distribution gain alone is compatible with tighter
adaptation to demonstrated paths.  Pairing the same frozen policy with
controlled camera, language, embodiment, and scene shifts tests whether the
learned action dependencies remain useful after contextual correlations change.
Cross-benchmark transfer, real-camera perturbations, and
intervention-based recovery tests extend this distinction to new task
distributions and control interfaces.  Reporting suite-resolved changes
alongside aggregate accuracy reveals where a regularizer transfers and where
its assumptions fail.

\subsection{Temporal Scope, Scaling, and Extensions}

CAP regularizes a length-\(H\) expert chunk.  Closed-loop behavior additionally
depends on distribution shift induced by policy rollouts, partial observability, controller latency,
error recovery, and the executed prefix \(e\).  The factorization family
\(\Pi_K\) determines which future relations receive supervision, while
\(H\), \(K\), and \(e\) determine how those relations align with physical
execution.  The \(K=0\) ablation retains the reference's average LIBERO
performance while trading Object success for gains on Goal and Long
(Table~\ref{tab:cap-controlled-sweep}).  This supports treating the anchor as
a tunable structural bias.  Unrestricted orders retain the chronological
LTR term, so the auxiliary prefix can be adjusted independently of deployment.
Longer chunks
provide more conditional subsets, larger anchors preserve more deployment
order, and shorter execution prefixes increase feedback.  Learning the reveal
distribution from task dynamics offers a direct extension of fixed uniform
sampling.  A learned proposal can condition reveal orders on the observation,
predicted uncertainty, or estimated contact phase.  Curriculum schedules can
progress from near-LTR orders to harder completion problems, and
latency-aware anchors can align the protected prefix with actions committed by
the controller.  These designs turn \(M\), \(K\), and \(\lambda\) from global
constants into properties of the current control problem.

Scaling introduces a second design axis: the representation layers reached by
the auxiliary gradient.  The decoder-only Goal experiment weakens the result,
suggesting that useful CAP constraints can shape the shared multimodal
context.  Layer-selective routing can test this interaction across frozen
backbones, adapters, resamplers, and action heads.  Applying the same controls
to broadly pretrained VLAs can then determine whether factorization
regularization complements additional data and capacity or changes its
preferred strength as scale increases.

The same principle applies whenever an action generator admits a base
deployment objective and valid subset-conditioned auxiliary objectives.
Autoregressive token policies can permute future token groups, diffusion and
flow policies can vary revealed coordinates and noise levels, and
multi-effector policies can study factorizations across time and coordinated
effectors.  Each extension requires a matched base-objective control and a
clear definition of the variables being permuted.  Diffusion and flow models
also introduce choices over clean-subset schedules, coordinate-wise noise,
denoising paths, and inference budgets.  Appendix \ref{app:permdiffusion}
provides a concrete diffusion construction.

%% file: figures/loss_overview.tex
\resizebox{\textwidth}{!}{%
\begin{tikzpicture}[
  font=\small,
  >=Latex,
  panel/.style={
    draw=black!35,
    dashed,
    rounded corners=4pt,
    inner sep=6pt
  },
  title/.style={font=\bfseries\large, align=center},
  label/.style={font=\bfseries, align=right},
  expert/.style={
    draw=blue!55!black,
    rounded corners=2pt,
    fill=blue!9,
    minimum width=7.4mm,
    minimum height=6.7mm,
    inner sep=1pt
  },
  prediction/.style={
    draw=orange!65!black,
    rounded corners=2pt,
    fill=orange!13,
    minimum width=7.4mm,
    minimum height=6.7mm,
    inner sep=1pt
  },
  reveal/.style={
    draw=violet!65!black,
    rounded corners=2pt,
    fill=violet!10,
    minimum width=7.4mm,
    minimum height=6.7mm,
    inner sep=1pt
  },
  anchortoken/.style={
    draw=green!45!black,
    very thick,
    rounded corners=2pt,
    fill=green!15,
    minimum width=7.4mm,
    minimum height=6.7mm,
    inner sep=1pt
  },
  loss/.style={
    draw=red!55!black,
    rounded corners=3pt,
    fill=red!6,
    align=center,
    minimum height=8.5mm,
    inner xsep=5pt
  },
  combine/.style={
    draw=black!65,
    very thick,
    rounded corners=3pt,
    fill=green!7,
    align=center,
    minimum height=9mm,
    inner xsep=7pt
  },
  flow/.style={-Latex, thick, draw=black!65},
  compare/.style={-Latex, thin, draw=red!65!black}
]

\node[title] (ltr-title) at (4.05,5.55)
  {Standard LTR objective\\[-1mm]\normalsize one chronological factorization};
\node[label, anchor=east] (ltr-expert-label) at (1.18,4.45)
  {expert\\targets};
\node[label, anchor=east] (ltr-policy-label) at (1.18,3.05)
  {policy\\outputs};

\foreach \i in {1,...,6} {
  \pgfmathsetmacro{\x}{1.65 + 1.02*(\i-1)}
  \node[expert] (le\i) at (\x,4.45) {$a_{\i}$};
  \node[prediction] (lp\i) at (\x,3.05) {$\hat a_{\i}$};
  \draw[compare] (le\i.south) --
    node[right, font=\scriptsize, text=red!65!black] {$\ell_{\i}$}
    (lp\i.north);
}
\foreach \i in {1,...,5} {
  \pgfmathtruncatemacro{\j}{\i+1}
  \draw[flow] (lp\i.east) -- (lp\j.west);
}
\node[font=\scriptsize, text=black!60] at (4.0,2.45)
  {teacher forcing reveals the same prefix at every slot};
\node[loss] (ltr-loss) at (4.0,1.63)
  {$\displaystyle \mathcal L_{\mathrm{LTR}}
    =\sum_{i=1}^{H}-\log p_\theta(a_i\mid\context,a_{<i})$};

\node[panel,
      fit=(ltr-title)(ltr-expert-label)(ltr-policy-label)
          (le1)(le6)(lp1)(lp6)(ltr-loss),
      inner xsep=11pt, inner ysep=8pt] (ltr-panel) {};

\node[title] (cap-title) at (12.65,5.55)
  {CAP regularizer\\[-1mm]\normalsize sampled reveal orders};
\node[label, anchor=east] (cap-expert-label) at (9.53,4.45)
  {expert\\chunk};
\foreach \i in {1,...,6} {
  \pgfmathsetmacro{\x}{10.02 + 1.12*(\i-1)}
  \node[expert] (re\i) at (\x,4.45) {$a_{\i}$};
}
\node[font=\scriptsize, text=black!60] at (12.78,3.88)
  {physical slots fixed; reveal order changes};

\node[label, anchor=east] at (9.53,3.18) {$z^{(1)}$};
\foreach \slot [count=\pos] in {1,4,2,6,3,5} {
  \pgfmathsetmacro{\x}{10.02 + 1.12*(\pos-1)}
  \ifnum\pos=1
    \node[anchortoken] (rA\pos) at (\x,3.18) {$a_{\slot}$};
  \else
    \node[reveal] (rA\pos) at (\x,3.18) {$a_{\slot}$};
  \fi
}
\foreach \i in {1,...,5} {
  \pgfmathtruncatemacro{\j}{\i+1}
  \draw[flow] (rA\i.east) -- (rA\j.west);
}

\node[label, anchor=east] at (9.53,2.28) {$z^{(2)}$};
\foreach \slot [count=\pos] in {1,3,6,2,5,4} {
  \pgfmathsetmacro{\x}{10.02 + 1.12*(\pos-1)}
  \ifnum\pos=1
    \node[anchortoken] (rB\pos) at (\x,2.28) {$a_{\slot}$};
  \else
    \node[reveal] (rB\pos) at (\x,2.28) {$a_{\slot}$};
  \fi
}
\foreach \i in {1,...,5} {
  \pgfmathtruncatemacro{\j}{\i+1}
  \draw[flow] (rB\i.east) -- (rB\j.west);
}
\node[font=\scriptsize, text=green!35!black, anchor=south]
  at (10.02,3.57) {$K{=}1$ anchor};

\node[loss, font=\footnotesize] (cap-loss) at (12.45,1.16)
  {$\displaystyle \mathcal L_{\mathrm{CAP}}
    =\frac{1}{M}\sum_{m=1}^{M}\sum_{j=1}^{H}
      -\log p_\theta(a_{z_j^{(m)}}\mid\context,a_{z_{<j}^{(m)}})$};

\node[panel,
      fit=(cap-title)(cap-expert-label)(re1)(re6)(rA1)(rA6)(rB1)(rB6)(cap-loss),
      inner xsep=11pt, inner ysep=8pt] (cap-panel) {};

\node[combine] (joint) at (8.25,-0.68)
  {$\displaystyle \mathcal L
    =(1-\lambda)\mathcal L_{\mathrm{LTR}}
     +\lambda\mathcal L_{\mathrm{CAP}}$
   \quad $\longrightarrow$ \quad shared VLA policy};
\node[combine, right=7mm of joint, fill=blue!6] (deploy)
  {deployment:\\deterministic LTR};
\draw[flow] (ltr-loss.south) -- ++(0,-3mm) -| ($(joint.north)+(-22mm,0)$);
\draw[flow] (cap-loss.south) -- ++(0,-3mm) -| ($(joint.north)+(22mm,0)$);
\draw[flow] (joint) -- (deploy);

\end{tikzpicture}%
}

%% file: sections/appendix_d_robot_visualization.tex
\section{Robot Rollout Visualization and Benchmark Diagnostics}
\label{app:robot-visualization}

We examine a stove-placement case under a LIBERO-Plus
\citep{fei2026liberoplus} table-appearance shift and two additional tasks
from original LIBERO \citep{liu2023libero} Long.  Full trajectories, task
predicates, and a demonstration audit reveal differences in subgoal completion
and post-success behavior that binary success rates obscure.

\subsection{Matched Rollouts Through the Time Limit}
\label{app:full-horizon-case}

We compare the matched LTR-1pass, LTR-2pass, and CAP experiments with
\(H=16,O=1,e=8\), 30k training updates, and seed zero.  CAP uses
\(M=1,K=1,\lambda=0.25\).  The three policies start from the same simulator
state in table-appearance variant 17, selected from historical evaluations
in which CAP succeeded and both LTR controls failed.  Each policy is executed
afresh through the full 520-action limit, with continued replanning after
first success and a frame recorded after every action.
Table~\ref{tab:full-horizon-case} reports these fresh diagnostic outcomes;
aggregate benchmark scores retain the standard first-success stopping protocol.

\begin{figure}[!htbp]
  \centering
  \includegraphics[width=\textwidth]{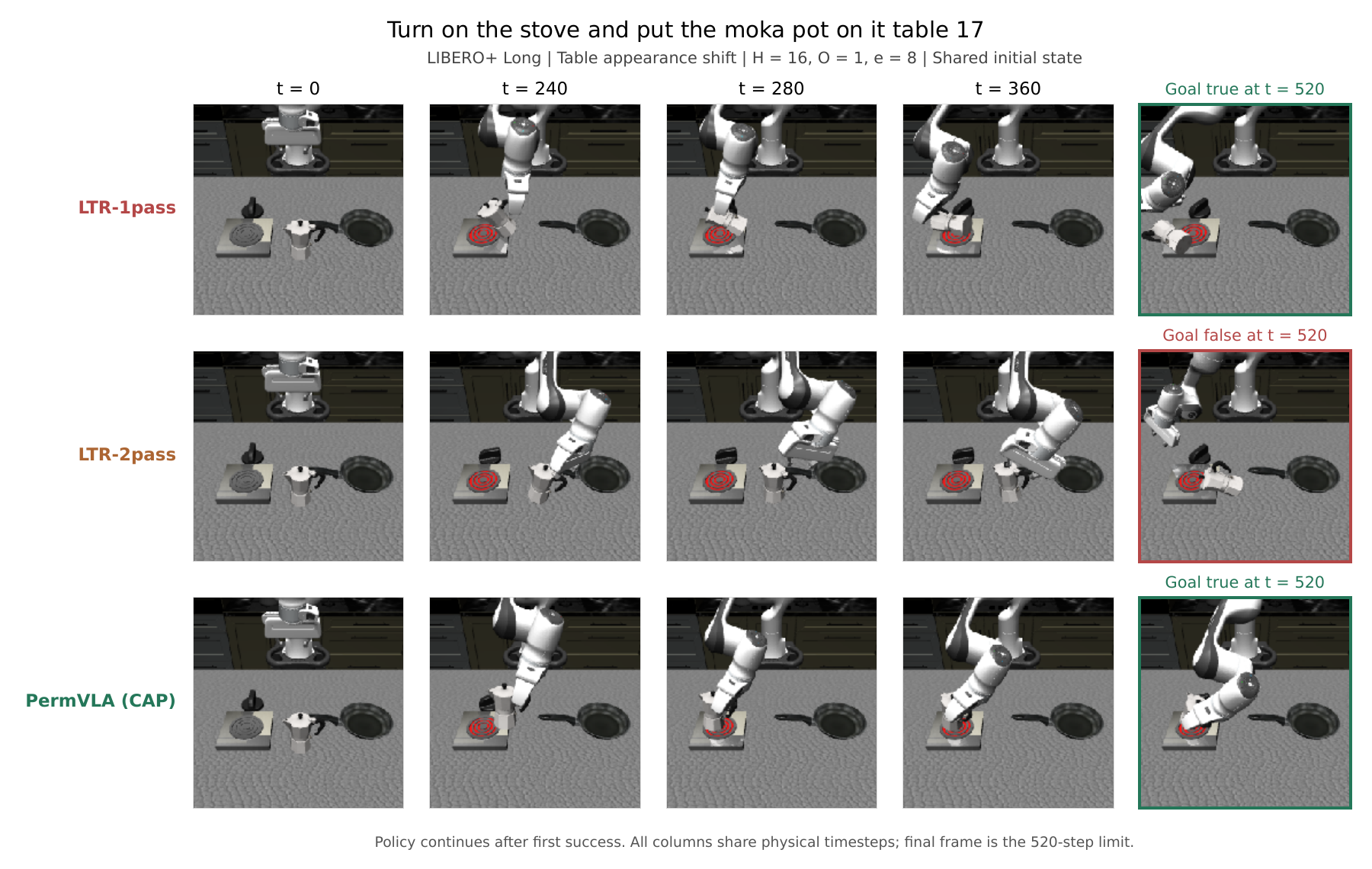}
  \caption{Matched robot rollouts under a table-appearance shift.  Columns
  align physical action steps across policies, including the initial state
  and the full time limit.  The final-frame border records the instantaneous
  official goal predicate.  Both LTR-1pass and CAP satisfy that predicate at
  the limit, despite visibly different pot poses and gripper interactions.}
  \label{fig:full-horizon-case}
\end{figure}

\begin{table}[!htbp]
  \centering
  \caption{Measurements for the full-horizon case in
  Figure~\ref{fig:full-horizon-case}.  Goal losses count transitions from true
  to false after first success.  Tilt is the angle between the pot's body
  vertical axis and the world vertical; contact denotes any pot--gripper
  contact at the limit.}
  \label{tab:full-horizon-case}
  \tablebodyfont
  \setlength{\tabcolsep}{5pt}
  \begin{tabular}{lccccc}
    \toprule
    \multicolumn{1}{c}{Policy} & First success & Goal at limit & Goal losses &
      Tilt at limit & Gripper contact \\
    \midrule
    LTR-1pass & 240 & True & 4 & \(90.0^\circ\) & Absent \\
    LTR-2pass & -- & False & 0 & \(58.4^\circ\) & Absent \\
    CAP & 265 & True & 0 & \(29.8^\circ\) & Present \\
    \bottomrule
  \end{tabular}
\end{table}

\begin{figure}[!htbp]
  \centering
  \includegraphics[width=0.94\textwidth]{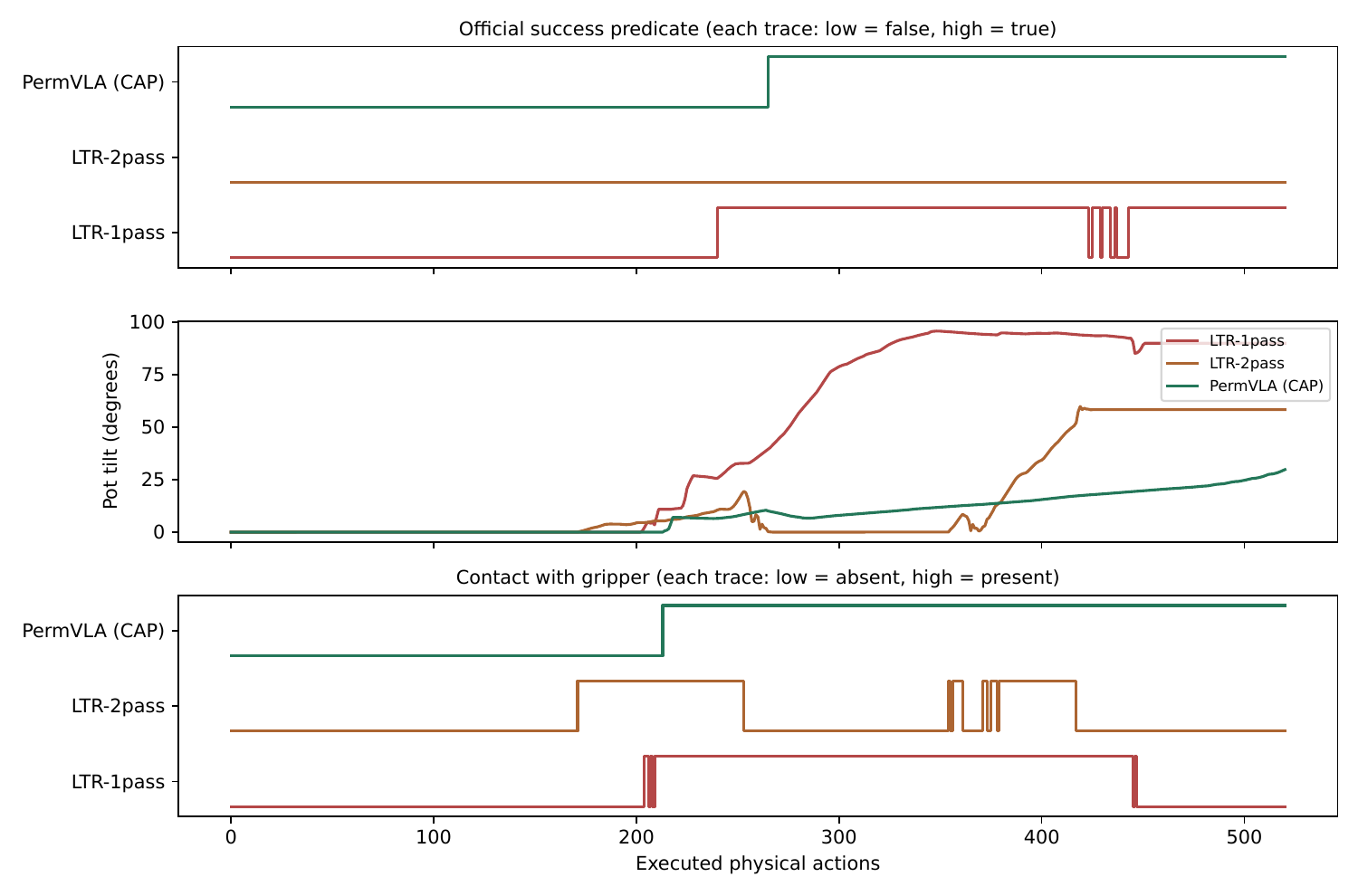}
  \caption{State diagnostics throughout the same rollouts.  The top and
  bottom panels vertically offset each policy's binary trace for readability.
  Success and gripper contact are evaluated from the simulator at each step;
  pot tilt is measured from its body orientation.  Sustained predicate
  satisfaction can coexist with a tilted pot or continued gripper contact.}
  \label{fig:full-horizon-diagnostics}
\end{figure}

CAP retains the successful predicate after its first occurrence and leaves
the pot closer to upright.  LTR-1pass repeatedly crosses the success boundary
and eventually leaves the pot on its side.  LTR-2pass contacts the stove
without satisfying the placement predicate.  CAP's final pose remains tilted
and in contact with the gripper: its continuation improves on the controls
while still falling short of upright, released placement.
Figure~\ref{fig:full-horizon-diagnostics} tracks these differences over time.

\subsection{What the Success Predicate Measures}
\label{app:placement-predicate}

For this task, the goal is the conjunction of \texttt{Turnon(stove)} and
\texttt{On(pot, cook region)}.  The placement test checks that the pot's body
origin lies inside a prescribed region above the cooking site and that the
pot contacts the stove.  Upright orientation, support through the pot's base,
gripper disengagement, and persistence over a settling interval are absent
from this conjunction.  Consequently, a pot resting on its side can satisfy
the same predicate as a correctly placed pot.  Stopping evaluation at the
first successful state also leaves subsequent loss of the goal unobserved.

The case reveals a gap between geometric task completion and the constraints
of reliable physical manipulation.  Binary success hides both the quality of
the final placement and the behavior after first success.  In particular,
LTR-1pass receives success credit for a sideways placement, while CAP receives
it with continuing gripper contact.  These policies were trained by imitation;
the observed mismatch concerns which behaviors the evaluation accepts.

\subsection{Demonstration Endings and Supervision Quality}
\label{app:demonstration-endings}

We audit all 41 demonstrations of this task in the processed LIBERO training
data used by the compared policies.  Of these, 38 end with an opening-gripper
command and three end with a closing command.  Among the 38 opening endings,
the final contiguous opening-command segment has a median length of nine
retained steps, with a range of 1--42; 36 contain at most 16 such steps.
These lengths count stored demonstration steps following the last switch to
an opening command.  Actual detachment, first task success, and elapsed
settling time are separate quantities.

The audit identifies heterogeneous terminal supervision: some trajectories
end with the gripper commanded closed, and most opening endings provide a
short segment for learning release and withdrawal.  Such demonstrations can
blur the distinction between reaching a goal region and completing a clean
placement.  We hypothesize that this ambiguity can encourage incomplete
finishing behavior during imitation learning, especially when a policy
continues beyond the states covered by the demonstrations.  CAP enriches the
conditional prediction problems built from these actions while retaining
the expert targets; improving those targets remains a complementary need.

\paragraph{Implications for benchmark design.}
This task motivates auditing demonstrations for consistent task endings and
recording release, arm withdrawal, and a settling interval as part of the
supervised trajectory.  Placement evaluation should include task-appropriate
orientation, stable support, gripper disengagement, and sustained completion.
Reporting these checks alongside success rates and OOD tests would distinguish
robust manipulation from transient or physically incomplete goal satisfaction.
Extending the audit across tasks would measure the prevalence of this mismatch.

\subsection{Additional Tasks in Original LIBERO}
\label{app:original-libero-cases}

\begin{figure}[!htbp]
  \centering
  \includegraphics[width=\textwidth]{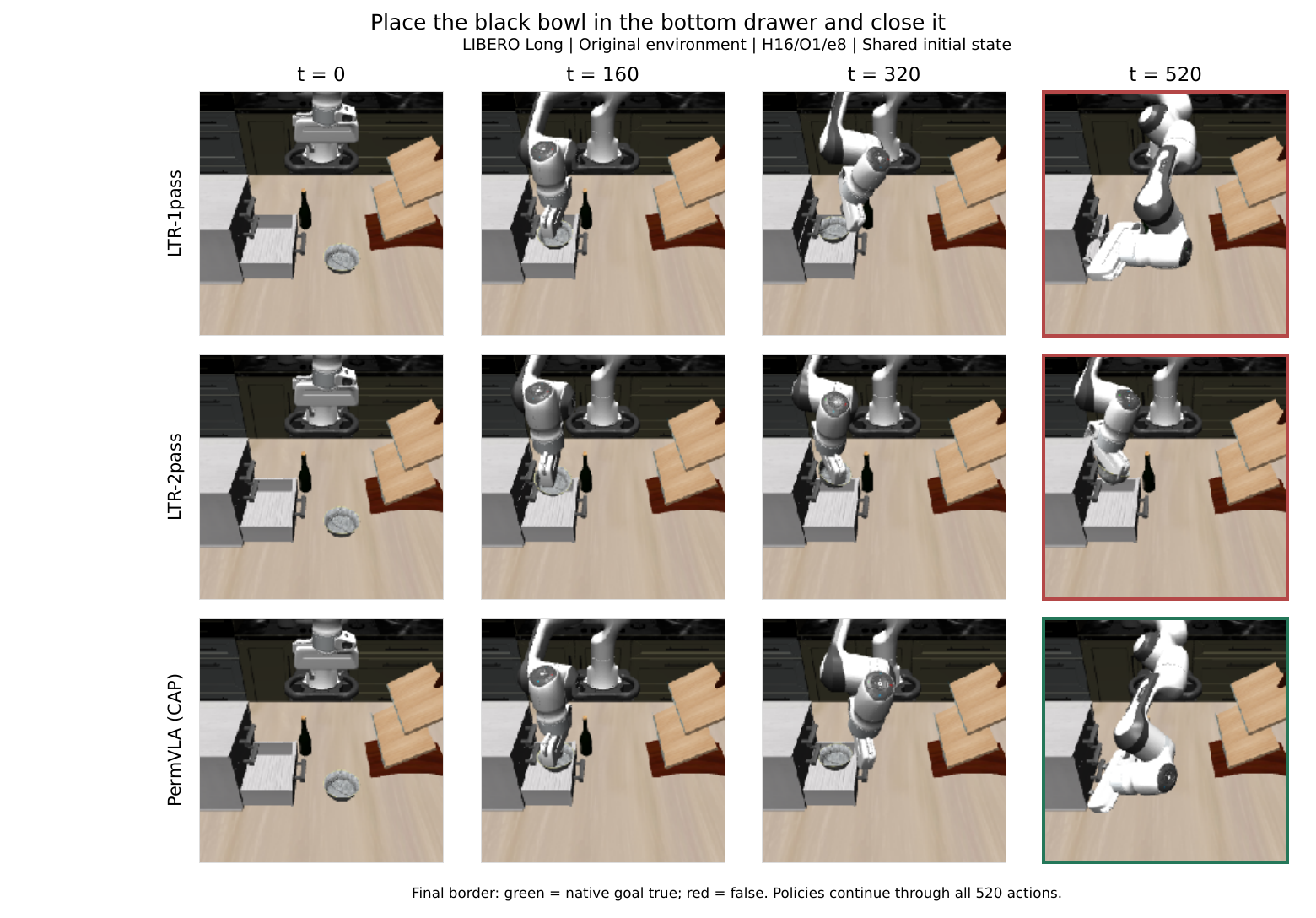}
  \caption{Bowl placement and drawer closure in original LIBERO.  CAP satisfies
  both subgoals from action 386 through the time limit.  At the limit,
  LTR-1pass satisfies bowl containment alone, while LTR-2pass satisfies
  neither containment nor closure.  Each row shows actual policy execution
  through all 520 actions.}
  \label{fig:original-bowl-drawer}
\end{figure}

\begin{figure}[!htbp]
  \centering
  \includegraphics[width=\textwidth]{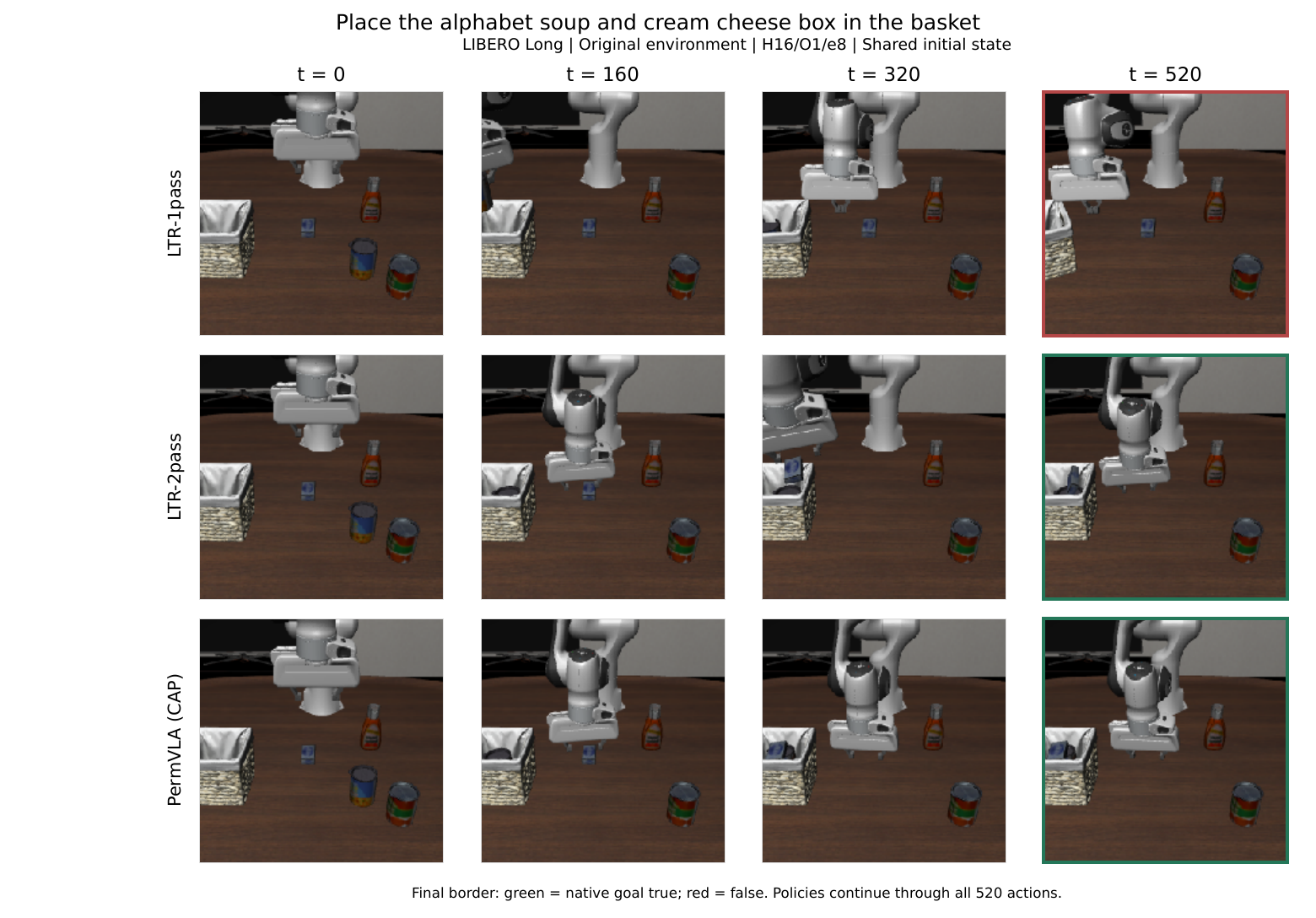}
  \caption{Two-object basket placement in original LIBERO.  CAP and LTR-2pass
  first satisfy the joint placement predicate at actions 230 and 338,
  respectively, and retain it through action 520.  LTR-1pass places the
  alphabet soup in the basket but leaves the cream cheese box outside.
  Common action-step columns expose progress through both placements.}
  \label{fig:original-two-object-basket}
\end{figure}

Figures~\ref{fig:original-bowl-drawer} and \ref{fig:original-two-object-basket}
illustrate successful CAP executions on two tasks from original LIBERO Long:
placing a bowl in a drawer and closing it, and placing two objects in a
basket.  We use the first evaluation initial state for each task and show
both matched LTR controls.  The matched policies and full-budget protocol
follow Section~\ref{app:full-horizon-case}.  Within each
task, the initial simulator state and camera image are identical across
policies.  Columns align physical action steps; a green or red final border
indicates whether the official goal predicate holds at the time limit.

The drawer and basket cases distinguish completing an individual placement
from satisfying the full instruction.  CAP completes both stages of the
drawer task, while LTR-1pass places the bowl inside but leaves the drawer
open.  In the basket task, CAP and LTR-2pass place both objects and maintain
the goal through the remaining actions; LTR-1pass completes only the soup
placement.  These sequences illustrate differences in multi-stage completion
under the same architecture and execution protocol.

